**Improving LLM Leaderboards with Psychometrical Methodology**

Denis Federiakin, Ph.D.;

Department of Business and Economics Education, Johannes Gutenberg University Mainz, Germany, denis.federiakin@uni-mainz.de;

Institute of Psychology, Goethe University Frankfurt, Germany, federiakin@psych.uni-frankfurt.de.

## Abstract

The rapid development of large language models (LLMs) has necessitated the creation of benchmarks to evaluate their performance. These benchmarks resemble human tests and surveys, as they consist of sets of questions designed to measure emergent properties in the cognitive behavior of these systems. However, unlike the well-defined traits and abilities studied in social sciences, the properties measured by these benchmarks are often vaguer and less rigorously defined. The most prominent benchmarks are often grouped into leaderboards for convenience, aggregating performance metrics and enabling comparisons between models. Unfortunately, these leaderboards typically rely on simplistic aggregation methods, such as taking the average score across benchmarks.

In this paper, we demonstrate the advantages of applying contemporary psychometric methodologies – originally developed for human tests and surveys – to improve the ranking of large language models on leaderboards. Using data from the Hugging Face Leaderboard as an example, we compare the results of the conventional naïve ranking approach with a psychometrically informed ranking. The findings highlight the benefits of adopting psychometric techniques for more robust and meaningful evaluation of LLM performance.

# 1. Introduction

Ever since the introduction of ChatGPT by OpenAI in the fall of 2022, Artificial Intelligence (AI)-enhanced chatbots based on Large Language Models (LLMs) have become game-changers in many areas of human activity, revolutionizing them across the board (e.g., Iu & Wong, 2023; Mehnen et al., 2023; Biswas, 2023). The potential of AI-assisted tools to facilitate and accelerate the execution of many professional and everyday-life functions is rooted in the cognitive capacities of LLMs to act as virtually universal assistants in information processing. This has also paved the way for the constant and rapid improvement of the cognitive performance of AI-assisted tools, supported by the steady development and release of different LLMs.

Correspondingly, the need for testing and comparing different LLMs has emerged. The necessity for evidence-based comparison of the capabilities of various LLMs has resulted in the rise of benchmarks – sets of tasks and questions provided to the models in natural (or visual) language, requiring LLMs to generate responses. These responses are then judged on their correctness, as the questions are presumed to have definitive answers (similar to human tests). As a result, thousands of specific benchmarks have been developed over the past few years (Guo et al., 2023; Chang et al., 2023). By now, benchmarks for LLMs exist in virtually every professional field, cognitive process, or aspect of ethics.

One such benchmark, MMLU (Hendrycks et al., 2020), has gained special popularity. MMLU contains 15,908 multiple-choice questions covering 57 different topics, ranging from high elementary mathematics to U.S. foreign policy. This breadth allows MMLU to serve as a proxy for assessing the general awareness of LLMs about the world – essentially a measure of their general knowledge. However, MMLU is far from the only popular benchmark.

With the diversification of various benchmarks, the problem of systematizing information on LLM performance has also arisen. With new benchmarks and LLMs appearing almost daily, the issue of comparing and integrating information from these benchmarks has become increasingly important. Correspondingly, multiple LLM leaderboards have emerged. These leaderboards openly publish information on how well various LLMs perform across a selected set of benchmarks. As a result, LLM leaderboards have become one of the most important and trustworthy sources of information on the relative capabilities of different LLMs.

One such leaderboard – the Hugging Face Leaderboard (Beeching et al., 2023; https://huggingface.co/spaces/HuggingFaceH4/open_llm_leaderboard) – is particularly significant due to its community support and popularity. This leaderboard has now become one of the gold standards for LLM comparison.

Yet, despite the widespread popularity and attention surrounding LLMs in general and LLM benchmarking in particular, there is a surprising lack of literature that examines the quality of the benchmarks being used (Wang et al., 2023) or the ranks provided by the leaderboards. Such quality can be analyzed using elements of psychometric methodology, which aims to ensure high-quality information for decision-making (e.g., ranking individuals based on some ability or trait). This methodology is responsible for the development of key psychological concepts such as IQ and the Big Five personality theory, having been carefully refined over the past century.

In general, psychometricians enhance various tests and surveys by refining item formulations and statistically analyzing their performance to ensure that the claims based on test results are valid and reliable. However, to the best of our knowledge, there are almost no studies on the quality of the prominent leaderboards. This paper aims to address this gap.

The primary purpose of this paper is to investigate the psychometric quality of the Hugging Face Leaderboard. We begin by discussing the differences between test development and benchmark development practices, highlighting the advantages and shortcomings of each. Next, we provide a brief introduction to psychometric modeling, which is used to establish validity evidence for human tests and surveys. Following this, we describe the methodology of the study.

We then apply this modeling approach to the performance data of various LLMs on Hugging Face Leaderboard v.1 (Study 1) and v.2 (Study 2). Subsequently, we compare the reported average scores from the leaderboards to the estimated factor scores obtained in Studies 1 and 2, and we discuss their differences (Study 3). Finally, we present a discussion on the future of psychometric analysis in evaluating LLM benchmarking data.

## 2. Test-development vs. benchmark-development

The entirety of the test development practice can be roughly summed up as proving that a test measures what it is intended to measure. This is referred to as providing validity evidence for the specific types of claims made about respondents based on the test results (AERA, APA, NCME, 2014). The focus on the types of claims is crucial, as a test developed for, say, research purposes might be unsuitable for use in clinical practice (Truijens et al., 2019). Hence, the purpose of test application (i.e., the targeted claim) is critical for constructing validity arguments.

There are numerous approaches to systematizing this process, but Evidence-Centered Design (ECD) is one of the most prominent in education, social sciences, and cognitive sciences (Riconscente et al., 2015; Mislevy et al., 2003; Mislevy & Haertel, 2006). The aim of ECD is to eliminate alternative explanations for the targeted claims. For instance, if testing results suggest that respondent A has lower science literacy than respondent B, test developers must demonstrate that:

1. The test results genuinely reflect science literacy (as defined by the test developers) and not, for example, mathematical ability, general intelligence, or any other construct.

2. The test results differ due to a systematic difference in science literacy between respondents, rather than due to randomness in responses (i.e., differences attributable to the standard error of measurement) or construct-irrelevant factors (such as demographic differences).

Each of these conclusions is based on numerous lower-level conclusions and studies, as addressing them requires unpacking a multitude of intertwined concepts and phenomena from social and cognitive science. As a result, ECD is a highly complex methodology that effectively integrates the test development process with the gathering of validity evidence. However, when properly implemented, it ensures that the test meets the Standards for Educational and Psychological Testing (AERA, APA, NCME, 2014). This means that it is clear what the test measures and how precise it is in accomplishing its intended purpose.

This approach essentially reflects the reasoning behind the *causal theory of measurement* (Markus & Borsboom, 2013; Mari et al., 2023). According to this perspective, when developing any measurement instrument, one must demonstrate that the results of the measurement are caused by the characteristic the instrument is intended to measure. Applied to test development using the ECD framework, this means that once psychometricians and test developers have ruled out competing explanations for the causes of the observed item responses, the only remaining explanation is the targeted characteristic.

This naturally brings up the issue of social constructivism (see Fried, 2017). Social constructivism essentially means that the studied phenomenon does not exist by itself, but the way it unfolds is defined by the approach taken to interact with it. In other words, the phenomenon is not "set in stone" but is rather built up (*constructed*) while it is studied. For example, researcher A can define mathematical ability as proficiency in addition and subtraction, while researcher B can define it as proficiency in division and multiplication. Then, the empirical consequences of this theoretical difference can arise to the point of their contradictions even though both constructs are called mathematical ability. This is a special type of problem in itself as it creates a number of theories that are seemingly similar, but describe different aspects of the same or maybe even different phenomena (Elson et al., 2023). However, this results in all tests with convincing validity evidence having a clear and detailed definitions of what they intend to measure – it is a side-product of test-developers agreeing on what they mean when they name the construct.

The development of benchmarks, however, follows another path. While not said explicitly in any of the papers published alongside the benchmarks, the authors, apparently, ground their philosophy of measurement in *representativism*. In social science measurements, representativism assumes that the items exhaust all aspects (or at least are a representative sample from all possible aspects) of the underlying quality that the test targets. For example, according to this perspective, the length of the table *is represented* by the reading on the ruler. At the same time, within the causal approach to measurement this idea would sound as "the length of the table *causes* readings on the ruler". While this might appear as a semantic wordplay, this slight philosophical difference has large practical consequences for the measurement design and mathematical modeling used for it (Markus & Borsboom, 2013).

This naturally raises the issue of social constructivism (see Fried, 2017). Social constructivism essentially posits that the phenomenon under study does not exist independently but is shaped by the approach taken to interact with it. In other words, the phenomenon is not "set in stone" but is instead *constructed* during the process of studying it. For example, researcher A might define mathematical ability as proficiency in addition and subtraction, while researcher B might define it as proficiency in division and multiplication. The empirical consequences of these differing theoretical definitions can lead to contradictions, even though both constructs are referred to as mathematical ability.

This is a unique problem because it creates multiple theories that appear similar but actually describe different aspects of the same – or potentially even different – phenomena (Elson et al., 2023). However, this issue has a silver lining: all tests with convincing validity evidence are accompanied by clear and detailed definitions of what they intend to measure. This is a natural byproduct of test developers agreeing on what they mean when they define a construct.

The development of benchmarks, however, follows a different trajectory. While this is not explicitly stated in the papers accompanying benchmarks, it appears that their authors ground their philosophy of measurement in *representativism*. In social science measurement, representativism assumes that the items in a test exhaust all aspects – or at least constitute a representative sample of all possible aspects – of the underlying quality the test aims to measure. For example, from this perspective, the length of a table is *represented* by the reading on a ruler.

In contrast, the causal approach to measurement would frame this differently, stating that "the length of the table *causes* the readings on the ruler". While this may seem like semantic wordplay, this subtle philosophical distinction has significant practical consequences for measurement design and the mathematical modeling used to support it (Markus & Borsboom, 2013).

This is not to say that one approach is better than the other; rather, each is suited to different measurement contexts. While the "length of the table" example does not fully illustrate the differences between the approaches, the measurement of more complex properties does. For instance, to measure a complex LLM's "skills" from the representativist paradigm, one would need an enormous number of questions in a benchmark to capture as many aspects of the skill as possible. This requirement is evident from the general tradition of benchmark development – they typically consist of an extensive number of questions. While administering such a large number of test items to human respondents is impractical, the infinite stamina of LLMs allows benchmarks to adhere to the representativist measurement tradition.

However, this approach poses challenges – not because the number of measures needs to be vast, but because it must be *sufficiently large* to fully reflect all facets of the targeted characteristic. Herein lies the problem: elusive characteristics such as "mathematical reasoning," "higher-order reasoning," or "intelligence" have infinitely many aspects since the number of situations in which these characteristics manifest is infinite. Consequently, no set of items, no matter how large, can be definitively proven to represent the entirety of the target characteristic. This limitation means that even extensive question sets can systematically omit critical aspects of the phenomena under investigation.

Precisely because of this issue, assessment in education, social, and cognitive sciences has shifted toward the causal theory of measurement. According to this theory, it is not the size of the observation set that matters but the cause of the outcomes. A clear construct definition, coupled with a solid theoretical justification for item development, defines the essence of the construct to be measured. This approach liberates the test developer from the need to represent all possible aspects of the targeted trait; instead, only those aspects with well-defined parameters need to be measured.

These philosophical differences dictate many operational distinctions in benchmark and test development. Tests developed using the causal approach generally exhibit much clearer construct definitions and theoretical frameworks. Only a handful of benchmarks can match the level of clarity in answering the question "What is being measured?" that social science tests provide (Wang et al., 2023; e.g., Fei et al., 2023; Kardanova et al., 2024). However, this approach to test development is significantly more labor-intensive than benchmark development. In benchmark development, one can simply clone an item numerous times – an operation that, in ECD terms, involves generating observable indicators of the measured characteristic across various item contexts.

Nonetheless, the radically representativist approach of benchmark development does offer certain advantages. For instance, the standard error of measurement becomes vanishingly small with sufficiently long tests (e.g., Linacre, 1995). Thus, reliability of measurement is generally not a concern in this domain.

However, another challenge in benchmark development is sample size. For example, if a custom LLM is tuned for a very specific task, its progress can be tracked against a custom benchmark, even if the sample size (i.e., the number of LLMs) is just one. In contrast, psychometric statistical modeling requires a larger sample size, typically involving multiple respondents. This fundamental requirement makes psychometric modeling generally infeasible in such cases. This is one of the reasons why psychometric modeling has not yet been applied to analyze the quality of benchmarks or leaderboards.

### 3. A brief summary of psychometric modeling

The essence of psychometric modeling, as applied in this paper, can be defined as the controllable dimensionality reduction of an item set to a comprehensible number of dimensions using statistical models whose assumptions and interpretations align with the theoretical framework of the characteristic being measured. This means that making a claim about respondents based on 60 items is much more complex than doing so based on a single variable (e.g., ability). To reduce a relatively large number of items to fewer dimensions, specialized statistical models are employed. A key feature of these models is that they are interpretable in the context of the theory underlying the construct. For example, if a construct is presumed to consist of several interconnected traits, the model should reduce the observed variables to exactly this number of traits. Furthermore, since items are often designed to measure only specific traits rather than all traits simultaneously, the dimension reduction process should account for this aspect. This reflects the confirmatory (or reflective) approach to psychometric modeling (Hanafiah, 2020).

It is important to note that psychometric modeling encompasses multiple paradigms. One paradigm that appears to dominate the field is parametric latent variable modeling, which assumes the existence of a latent variable underlying several observed variables (Cai, 2012). Psychometric models in this paradigm link the observed variables (item responses) to the latent variable (e.g., ability) and are used to derive individual estimates of this latent variable.

However, this is by no means the only paradigm employed by psychometricians. Depending on the type of data, the intended claims about respondents, the purpose of the modeling, and the resources available, other paradigms can be applied.

These include non-parametric modeling of latent variables (Sijtsma & Van Der Ark, 2022), Classical Test Theory (CTT, Crocker & Algina, 1986; which seems to dominate the LLM benchmarking field), Generalizability Theory (Jiang, 2018), Network Modeling (Marsman et al., 2018; Costantini et al., 2015), and others. Notably, while CTT uses observed scores (e.g., item sums or averages), it is still based on statistical modeling and is not free from statistical assumptions (Novick, 1966). In this regard, parametric modeling of latent variables is not inherently "better" than CTT; rather, it is suited to solving certain problems and addressing specific questions that CTT cannot (Hambleton & Jones, 1993).

Here, we will elaborate on just two of these advantages: advanced ability estimation and test quality analysis. Simple averaging of responses to different test items assumes that all items are interchangeable. In other words, it assumes that all items have the same level of difficulty – a typical assumption in many equations of Classical Test Theory (CTT). This assumption, however, is almost never true. Psychometric latent variable models account for this by recognizing that correct responses to more difficult items should "count more" than responses to easier items. Furthermore, these models typically assume that items differ in their sensitivity to the latent variable. In other words, different items vary in their usefulness for estimating the targeted ability – some provide a lot of (Fisher) information (Muraki, 1993) about the ability, while others provide less. This approach allows psychometric models to account for differences not only across respondents but also across items, by describing items in terms of several properties (parameters). This enables the identification and exclusion of items that are nearly useless or even detrimental to ability estimation, such as items where the probability of a correct response decreases as ability increases.

The first major benefit of the most popular psychometric models is that they enhance the precision of the scale used to reflect ability. Because items differ in difficulty, the average (or sum) score across items operates on an ordinal scale (Stevens, 1946), as it reflects the number of items solved correctly, rather than the underlying cause of the responses. By contrast, one of the most widely used psychometric models, Item Response Theory (IRT), assumes that responding to an item correctly is a random event with a probability that depends on the latent variable (ability). Consequently, the estimates of latent ability derived from IRT are placed on an interval (metric) scale, defined by a logit-transformed probability scale. This provides the estimates with a unit of measurement, significantly enhancing the range of analyses that can be performed on this scale and improving the quality of respondent rankings.

This improvement, however, comes at the cost of several parametric assumptions required by these models. While the use of Stevens' (1946) scale classification is generally criticized by contemporary psychometricians (Zumbo & Kroc, 2019; Thomas, 2019), it remains helpful for understanding the comparative advantages and disadvantages of different approaches. Additionally, such advanced ability estimates filter out item-specific noise, allowing for the precise ranking of respondents based solely on variance common across items.

There are several families of models for parametric modeling of latent variables. For example, Item Response Theory (IRT; Van der Linden, 2018) maps discrete item responses to continuous latent traits, while Factor Analysis (FA; Brown, 2015) maps continuous responses to continuous latent traits. Psychometrics also employs specialized types of finite mixture modeling: Latent Class Analysis (Eshima, 2022) and Cognitive Diagnostic Modeling (von Davier & Lee, 2019) are used to map discrete responses to discrete latent characteristics, while models such as Latent Profile Analysis (Oberski, 2016) map continuous responses to discrete latent characteristics. Each of these families varies further in terms of their preferred estimation techniques.

Over the years, psychometricians have developed and routinely employed a variety of estimation methods, including Maximum Likelihood estimators (Chen & Zhang, 2021), Least Squares estimators (Savalei & Rosseel, 2022), Bayesian samplers (Levy & Mislevy, 2017; Wu et al., 2020), and regularization techniques (Robitzsch, 2023), as well as numerous modifications and combinations tailored to specific models. More recently, backpropagation has been proposed as a technique for estimating psychometric models (Urban & Bauer, 2021; Converse, 2021).

Particular attention, however, is given by psychometricians to global (model-level; Goretzko et al., 2024; Cai & Monroe, 2014) and local (item- and person-level; Köhler et al., 2020; Chalmers & Ng, 2017; Müller, 2020) model fit analysis. These procedures are crucial for verifying whether the theoretical assumptions underlying model development hold true in the observed data. Each branch of psychometric modeling contains countless models that differ in their assumptions about the data. Psychometricians continually strive to strengthen the connection between theoretical assumptions derived from educational, cognitive, and social sciences and the assumptions made by mathematical models.

Further elaboration on the broader purposes of psychometrics is beyond the scope of this paper.

## 4. Methods and Data

### 4.1 Analysis methodolody

In the case of applying psychometric models to LLM benchmarking data, there is a significant challenge of having more parameters than observations. Psychometric models estimate at least one parameter per observed variable (benchmark question) in Rasch (1960) models, and typically two or more in most other models. However, given the enormous number of benchmarking questions, there are not enough LLMs in existence to ensure that the number of observations (LLMs) exceeds the number of model parameters.

A relatively simple approach to addressing this problem is parceling – collapsing groups of observed variables into a single variable (Matsunaga, 2008) through summation or averaging. While this practice has its limitations (Little et al., 2002), the parceled information is precisely what the Leaderboard provides: it reports the average accuracy of LLMs responding to relatively homogeneous groups of items (i.e., averages calculated within each benchmark). The relative homogeneity of these items allows for a better understanding of relationships between groups and resolves the issue of having more model parameters than observations.

Parceling, however, introduces a less obvious challenge for the application of psychometric modeling, as parcels (especially those derived through averaging) produce continuous variables. On the one hand, this makes the data suitable for Factor Analysis (FA), but unsuitable for Item Response Theory (IRT). However, FA assumes that the observed variables follow a normal distribution, implying that they are not only continuous but also unbounded. By contrast, parcels derived from benchmark data have fixed lower and upper bounds of 0 and 1, respectively.

Fortunately, Samejima (1973, 1974) developed a unidimensional IRT model for such observed variables as a limiting case of her Graded Response Model for polytomous responses (Samejima, 1969), where the number of categories approaches infinity. Later, Wang and Zeng (1998) extended Samejima's model by introducing a parcel "difficulty" parameter. Subsequently, Ferrando (2002) formally explored the relationships between Wang and Zeng's modification of Samejima's model and linear FA. In this section, we will follow Ferrando's derivations.

Assume the observed data contains the performance of $M$ LLMs on $P$ observed variables (parcels). The observed performance $U_{mp}$ of model $m$ ($m = 1,2, \ldots, M$) on parcel $p$ ($p = 1,2, \ldots, P$) is standardized such that $0 < U_{mp} < 1$. Then, the CRM assumes

$$U_{mp} = logit^{-1}(V_{mp}) = logit^{-1}(\mu_p + \lambda_p\theta_m + \omega_{mp}), \qquad (1)$$

where $\mu_p$ is the easiness of parcel $p$ ($\boldsymbol{\mu} \in \mathbb{R}^P$),

$\lambda_p$ is the sensitivity of parcel $p$ to the changes in the target ability (the discrimination) ($\boldsymbol{\lambda} \in \mathbb{R}^P$, however, but typically $\boldsymbol{\lambda} \in \mathbb{R}_+^P$ for model identification),

$\theta_m$ is the value of latent variable denoting the target ability of model $m$ ($\theta \in \mathbb{R}$ in the unidimensional case; $\theta \sim \mathcal{N}(0,1)$ for model identification), and

$\omega_{mp}$ is the residual interaction of model $m$ and parcel $p$ ($\boldsymbol{\omega} \in \mathbb{R}^P$ ; $\omega_p \sim \mathcal{N}(0,\sigma_p)$).

Given that the distribution of $\boldsymbol{V} = \left(V_1, V_2, ..., V_p, ..., V_P\right)^T$ is assumed to be multivariate normal ($\boldsymbol{V} \sim \mathcal{N}(\boldsymbol{k}, \boldsymbol{S})$), the conditional distribution of $U_p$ is assumed to be

$$f(U_p|\theta_m) = \frac{1}{\sigma_p\sqrt{2\pi}} \frac{1}{U_p(1-U_p)} \exp\left\{-\frac{1}{2}\left[\frac{logit(U_p) - (\mu_p + \lambda_p\theta_m)}{\sigma_p}\right]^2\right\}. \qquad (2)$$

As noted by Ferrando (2002), Equation 2 is known as the $S_B$ distribution (Johnson, 1949) or the four-parameter log-normal distribution (Aitchison, & Brown, 1957). After the introduction of factor loading $\alpha_i = \frac{\lambda_p}{\sigma_p}$ and intercept $\tau_p = -\frac{\mu_p}{\lambda_p}$, Eq. (1) becomes

$$f(U_p|\theta_m) = \frac{\alpha_i}{\lambda_p\sqrt{2\pi}} \frac{1}{U_p(1-U_p)} \exp\left\{-\frac{1}{2}\left[\alpha_i\left\{\theta_m - \tau_p - \frac{logit(U_p)}{\lambda_p}\right\}\right]^2\right\}, \qquad (3)$$

Which is the form of CRM derived by Wang and Zeng (1998) – with the addition of the parcel intercept $\tau_p$. Thanks to the assumption of multivariate normality of $\boldsymbol{V}_p$, Equation 2 has the following conditional expectation:

$$E(U_p|\theta) = \int_{-\infty}^{+\infty} \frac{1}{1 + \exp[-(\mu_p + \lambda_p\theta + z\sigma_p)]} \varphi(z)\,dz, \qquad (4)$$

where $\varphi(z)$ is the density function of the standard normal distribution. This conditional expectation serves as Item Characteristic Curve of $U_p$.

The equivalence of Equation 2 to $S_B$ distribution (Johnson, 1949) is useful, as it highlights several desirable properties of this conditional distribution. Among them is the introduction of the skewness of the distribution density in the direction of 0.5 and the reduction in its variance proportionally to the proximity of the boundary value. Additionally, if $\mu_p = \theta_m$, the conditional expectation of Equation 4 is 0.5, further building analogies with more traditional IRT models, such as Rasch or 2PL models.

The key insight from these derivations is that the CRM assumes logistic link-function between $\boldsymbol{V}$ and $\boldsymbol{U}$, while the traditional linear FA assumes an identity link-function. This means that CRM can be approximated by an FA model on $\boldsymbol{V} = logit(\boldsymbol{U})$. However, in this approximation, the mean structure must be estimated alongside the covariance structure in the FA model. This process is referred to as the heuristic estimation procedure for CRM (Bejar, 1977).

One important detail is that the traditional CRM (Wang & Zeng, 1998) is estimated via Marginal Maximum Likelihood (Bock & Aitkin, 1981), where model-specific parameters ($\theta_m$) are marginalized to the standard normal distribution. In the realm of FA, this approach is known as full-information FA (Bartholomew, 1981).

The reason for resorting to the FA approximation of CRM is that it offers a significant advantage over the classical CRM parameterization: it provides a highly flexible set of tools for exploring the dimensionality of the latent factor space and enables advanced model fit analyses in a convenient manner. For example, a researcher can easily test whether multiple latent factors, rather than a single factor, are sufficient to approximate $\boldsymbol{S}$. Additionally, a researcher can explore alternative item loadings on latent factors.

In general, an FA model with a mean structure assumes that

$$\boldsymbol{V} = \boldsymbol{\mu} + \boldsymbol{\Lambda}\boldsymbol{\theta} + \boldsymbol{\varepsilon}, \tag{5}$$

where $\boldsymbol{\mu} \in \mathbb{R}^P$ is the vector of means of the observed variables (easiness from Eq. 1),

$\boldsymbol{\Lambda} \in \mathbb{R}^{P \times F}$ is matrix of factor loadings on $F$ factors ($F < P$; typically, for model identification $\boldsymbol{\Lambda} \in \mathbb{R}_+^{P \times F}$),

$\boldsymbol{\theta} \in \mathbb{R}^F$ is the vector of $F$ factors scores ($\boldsymbol{\theta} \sim \mathcal{N}(\boldsymbol{0}, \boldsymbol{\Xi})$ with $\text{diag}(\boldsymbol{\Xi}) = \boldsymbol{1}$ for model identification),

$\boldsymbol{\varepsilon} \in \mathbb{R}^P$ is the observed residuals vector ($\boldsymbol{\varepsilon} \sim \mathcal{N}(\boldsymbol{0}, \boldsymbol{\Psi})$).

FA attempts to approximate the sample variance-covariance matrix $\boldsymbol{S}$ from $\boldsymbol{V} \sim \mathcal{N}(\boldsymbol{k}, \boldsymbol{S})$ with the model-implied variance-covariance matrix $\boldsymbol{\Sigma}$:

$$\boldsymbol{S} \approx \boldsymbol{\Sigma} = \boldsymbol{\Lambda}\boldsymbol{\Xi}\boldsymbol{\Lambda}^T + \boldsymbol{\Psi}. \tag{6}$$

The log-likelihood of the data is the given by

$$\ell(\boldsymbol{\mu}, \boldsymbol{\Lambda}, \boldsymbol{\Xi}, \boldsymbol{\Psi}) = -\frac{M}{2}\left(P\log(2\pi) + \log|\boldsymbol{\Sigma}| + \text{trace}(\boldsymbol{S}\boldsymbol{\Sigma}^{-1}) + (\boldsymbol{\mu} - \boldsymbol{k})^T\boldsymbol{\Sigma}^{-1}(\boldsymbol{\mu} - \boldsymbol{k})\right), \tag{7}$$

where $|\boldsymbol{\Sigma}|$ is the determinant of $\boldsymbol{\Sigma}$. And the maximum likelihood estimates are given by

$$\left(\widehat{\boldsymbol{\mu}}, \widehat{\boldsymbol{\Lambda}}, \widehat{\boldsymbol{\Xi}}, \widehat{\boldsymbol{\Psi}}\right) = \arg\max_{\boldsymbol{\mu}, \boldsymbol{\Lambda}, \boldsymbol{\Xi}, \boldsymbol{\Psi}} \ell(\boldsymbol{\mu}, \boldsymbol{\Lambda}, \boldsymbol{\Xi}, \boldsymbol{\Psi}). \tag{8}$$

If specific constraints are imposed on $\boldsymbol{\Lambda}$, $\boldsymbol{\Xi}$, and $\boldsymbol{\Psi}$ (e.g., if a researcher assumes that certain items do not load on specific factors – then, corresponding elements of $\boldsymbol{\Lambda}$ being constrained to 0; or if there is an assumption of a single factor underlying all observed variables), then the model is said to be of a Confirmatory FA (CFA) nature.

If there are no constraints on these matrices, then the model is said to be of Exploratory FA (EFA) nature, with $\Xi = I$ (identity matrix, for identification purposes). While EFA tends to yield a better representation of $S$, CFA is generally preferred for higher-stakes decision-making, as it aligns with reflective (rather than formative) measurement principles (Hanafiah, 2020).

In CFA, the residual covariance matrix $\Psi$ is typically diagonal initially (with off-diagonal elements constrained to 0 and the main diagonal values estimated). Later, some constraints on the absent residual covariances can be relaxed based on automatic diagnostic such as model modification indices (Whittaker, 2012). If some observed variables in $S$ are more strongly correlated than $\Sigma$ predicts from $\Lambda\Xi\Lambda^T$, the model will be poor and adding estimated parameters in $\Psi$ becomes warranted.

However, such model modifications are also interpretationally significant. They reveal correlations in the data that are not explained by the factor scores. In such cases, the correlated observed variables should not be treated as locally independent by the model (conditionally independent given person parameters), as they reflect a substantial amount of common variance beyond that attributable to the latent factors ("common cause" for the observed variables). Therefore, interpreting residual covariances – or increasing the number of latent factors – requires careful content-based interpretation.

This modeling approach can be seeing as analogous to a variational autoencoder with the shallow decoder (one neuron in depth; Urban & Bauer, 2021). In this analogy, the neuron parameters from the (final) decoder layer correspond to item parameters, and the means of the observation-specific latent representation distributions serve as factor scores. However, while analyzing latent correlations between factors is not typically a focus in autoencoders, it is a central area of interest in FA. Additionally, autoencoders do not examine residual covariances. Furthermore, in CFA, the matrix of factor loadings $\Lambda$ includes some entries constrained to 0, whereas in autoencoders, all entries are estimated (making them more similar to EFA).

Alternatively, FA and IRT can be viewed as collaborative filtering engines (Bergner et al., 2022) designed to mathematically separate item (variable) parameters from person (observation) parameters.

In the field of CFA, several model fit indices are widely used:

- Root Mean Square Error of Approximation (RMSEA; Steiger, 1990; Steiger, 1998) as a measure of the average difference between the observed variance-covariance matrix $\boldsymbol{S}$ and the model-implied variance-covariance matrix $\boldsymbol{\Sigma}$ per degree of freedom,

- Standardized Root Mean Square Residual (SRMS; Hu & Bentler, 1998) as a measure of the averaged squared differences between each bivariate empirical correlation (in $\boldsymbol{S}$) and its corresponding model-implied counterpart (in $\boldsymbol{\Sigma}$),

- Comparative Fit Index (CFI; Bentler, 1990) as a measure of the relative improvement in model fit from the baseline model to the tested model,

- Tucker-Lewis Index (TLI; Tucker & Lewis, 1973) as a measure of the relative reduction in misfit from the baseline model to the tested model per degree of freedom.

For RMSEA and SRMS, the following cut-off criteria are applied: $0 <$ good fit $< 0.05 \leq$ acceptable fit $< 0.08 \leq$ poor fit. For CFI and TLI, the following cut-off criteria are applied: poor fit $\leq 0.9 <$ acceptable fit $\leq 0.95 <$ good fit $< 1$. Additionally, we report the $\chi^2$-statistic ($-2\ell(\boldsymbol{\mu}, \boldsymbol{\Lambda}, \boldsymbol{\Xi}, \boldsymbol{\Psi})$) which analyzes the statistical significance of the differences between $\boldsymbol{S}$ and $\boldsymbol{\Sigma}$ in the tested model. However, this statistic is not used for decision-making due to its generally high Type I error rates.

Nonetheless, this approach relies heavily on the assumption of multivariate normality of the observed data, which is rarely true. To address this, modifications such as the so-called Maximum Likelihood Robust (MLR) method have been developed. These modifications adjust the standard errors and the $\chi^2$-statistic value. Specifically, they scale the $\chi^2$-statistic by a factor determined by the multivariate skewness and kurtosis of the observed data (Yuan & Bentler, 2000) and estimate standard errors using a sandwich approach with the observed Fisher information matrix at the standard maximum likelihood estimates (Huber, 1967).

Once the model has been calibrated, the FA-reliability of measurement can be calculated. One common method is to compute the composite reliability coefficient (Bentler, 1968; often referred to as McDonald's $\omega$; McDonald, 1970) coefficient. In the case of a unidimensional model ($\theta \sim \mathcal{N}(0,1)$; regardless of presence or absence of residual covariances), it can be estimated as:

$$\rho = \frac{\left(\sum_{p=1}^{P} \lambda_p\right)^2}{\mathbf{1}^T \boldsymbol{S} \mathbf{1}}, \tag{9}$$

where the column-vector $\mathbf{1}$ of length $P$ is used to sum the entries in $\boldsymbol{S}$.

While the exact definition and interpretation of reliability remain a topic of debate (Cho, 2021), it is generally understood operationally as a measure of the non-randomness in observed variables, given the psychometric model being used. The closer $\rho$ is to 1 ($0 \leq \rho \leq 1$), the more reliable the results are, and the less the Standard Error of Measurement (S.E.) is.

The latent ability estimates can then be obtained via the Maximum a Posteriori (MAP) method (also known as Modal a Posteriori, Empirical Bayes, Bayes Modal, or naïve regression factor scores; Skrondal & Rabe-Hesketh, 2004). In this approach, the posterior distribution of the ability ($\boldsymbol{\theta}_m$) is defined as $P(\boldsymbol{\theta}_m | V_m) \propto P(V_m | \boldsymbol{\theta}_m) P(\boldsymbol{\theta}_m)$, i.e., as the product of the likelihood ($V_m | \boldsymbol{\theta}_m \sim \mathcal{N}(\boldsymbol{\mu} + \boldsymbol{\Lambda}\boldsymbol{\theta}_m, \boldsymbol{\Psi})$) and the priors ($\boldsymbol{\theta}_m \sim \mathcal{N}(\mathbf{0}, \boldsymbol{\Xi})$). The negative log-likelihood (up to constants) is:

$$-\ell(V_m | \boldsymbol{\theta}_m) = \frac{1}{2}(V_m - \boldsymbol{\mu} - \boldsymbol{\Lambda}\boldsymbol{\theta}_m)^T \boldsymbol{\Psi}^{-1}(V_m - \boldsymbol{\mu} - \boldsymbol{\Lambda}\boldsymbol{\theta}_m), \tag{10}$$

and the negative log-priors (up to constants) is:

$$-\log P(\boldsymbol{\theta}_m) = \frac{1}{2}\boldsymbol{\theta}_m^T \boldsymbol{\Xi}^{-1}\boldsymbol{\theta}_m. \tag{11}$$

The estimate is then obtained by minimizing the negative log-posterior:

$$\widehat{\boldsymbol{\theta}}_m = \arg\min_{\boldsymbol{\theta}_m}\left\{\frac{1}{2}\left[(V_m - \boldsymbol{\mu} - \boldsymbol{\Lambda}\boldsymbol{\theta}_m)^T \boldsymbol{\Psi}^{-1}(V_m - \boldsymbol{\mu} - \boldsymbol{\Lambda}\boldsymbol{\theta}_m) + \boldsymbol{\theta}_m^T \boldsymbol{\Xi}^{-1}\boldsymbol{\theta}_m\right]\right\}. \tag{12}$$

After taking the gradient of the minimized term in Equation 12 and setting it to 0, a closed-form solution for the factor scores can then be derived as:

$$\widehat{\boldsymbol{\theta}}_m = (\boldsymbol{\Lambda}^T \boldsymbol{\Psi}^{-1}\boldsymbol{\Lambda} + \boldsymbol{\Xi}^{-1})^{-1}\boldsymbol{\Lambda}^T \boldsymbol{\Psi}^{-1}(V_m - \boldsymbol{\mu}) = \boldsymbol{\Xi}\boldsymbol{\Lambda}^T(\boldsymbol{\Lambda}\boldsymbol{\Xi}\boldsymbol{\Lambda}^T + \boldsymbol{\Psi})^{-1}(V_m - \boldsymbol{\mu}). \tag{13}$$

Such factor scores provide more stable results, filtered out of noise, on the continuous unbounded scale, which offers a better metric than a simple average.

In addition to discussing and using the MAP point estimates of factor scores, their posterior variance is also critical. This variance serves as a measure of uncertainty (standard error) around the ability estimates and can be computed as:

$$S.E.\left(\widehat{\boldsymbol{\theta}}\right)^2 = var\left(\widehat{\boldsymbol{\theta}} | V_m\right) = (\boldsymbol{\Xi}^{-1} - \boldsymbol{\Lambda}^T \boldsymbol{\Psi}^{-1}\boldsymbol{\Lambda})^{-1} = \boldsymbol{\Xi} - \boldsymbol{\Xi}\boldsymbol{\Lambda}^T(\boldsymbol{\Lambda}\boldsymbol{\Xi}\boldsymbol{\Lambda}^T + \boldsymbol{\Psi})^{-1}\boldsymbol{\Lambda}\boldsymbol{\Xi}. \tag{14}$$

Note, that in Equation 14, the posterior variance estimate is constant and does not depend on $\widehat{\boldsymbol{\theta}}$. While this holds true in the case of CFA (since linear models provide a constant amount of Fisher information about the parameter), it is not true in the general case of IRT. IRT leverages non-linear models, and as a result, it explicitly accounts for the fact that different ability levels are associated with varying degrees of measurement precision.

The exact distribution of the Fisher information function depends on the properties of the item bank. However, it is generally the case that extreme (high and low) ability levels are measured with greater uncertainty, as item information is typically concentrated around medium levels of ability (the region of the highest density of the observations).

All analyses were conducted using the lavaan package (v. 0.6-17; Rosseel et al., 2023) for the statistical programming language R (v. 4.3.0). FA reliability was estimated using the semTools package (v. 0.5-6; Jorgensen et al., 2022).

## 4.2 Data

The data was retrieved from the Hugging Face Leaderboard (Beeching et al., 2023), where the performance of various models is published in open access. On this leaderboard, the evaluation results are presented in a few-shot manner (Brown et al., 2020). This means that the model is provided with a few examples of similar questions along with their correct answers before being evaluated on the target question. Within each task (type of questions within a benchmark), the number of shots is standardized. For some of the benchmarks, the performance is evaluated in a zero-shot manner, which is a special case of the few-shot approach where the number of preliminary examples administered is 0. All questions used in the leaderboard evaluations are Multiple-Choice (MC) with several options to choose from.

*Study 1* aimed to analyze the first version of the Hugging Face Leaderboard. Here we use the dataset retrieved from https://huggingface.co/spaces/open-llm-leaderboard-old/open_llm_leaderboard on November 30, 2024. This dataset contains the parceled performance of 3,792 LLMs on six benchmarks. These benchmarks include:

- AI2 Reasoning Challenge (ARC; Clark et al., 2018) – a set of 2,590 grade-school science questions designed to test commonsense knowledge and advanced methods for deeper text comprehension. These questions are administered in a 25-shot manner.

- HellaSwag (Zellers et al., 2019) – a set of 70,000 commonsense natural language inference MC questions. Evaluation questions are administered in a 10-shot manner.

- MMLU (see the introduction section above).

- TruthfulQA (Lin et al., 2021) – a set of 684 MC questions from 38 topics, including health, law, finance, and politics. This benchmark measures how well LLMs answer questions that some humans would answer incorrectly due to false beliefs or misconceptions. These questions are administered in a 6-shot manner.

- WinoGrande (Sakaguchi et al., 2021) – a set of 1,767 MC questions based on the Winograd Schema Challenge (Levesque et al., 2012), designed to measure commonsense reasoning from in-sentence context. These questions are administered in a 5-shot manner.

- GSM8K (Cobbe et al., 2021) – a set of 8,500 grade-school math questions and natural language solutions, used to probe the informal reasoning abilities of large language models. These questions are administered in a 5-shot manner.

*Study 2* aimed to analyze the second version of the Hugging Face Leaderboard. For this purpose, we used a dataset retrieved from https://huggingface.co/spaces/open-llm-leaderboard/open_llm_leaderboard on November 30, 2024. This dataset contains the parceled performance of 1,543 LLMs on six benchmarks:

- Instruction-Following Evaluation (IFEval; Zhou et al., 2024) – a set of approximately 500 items designed to evaluate LLMs' ability to follow 25 types of explicit, verifiable instructions. Examples include: "In your response, the word {word} should appear {N} times", "Finish your response with this exact phrase: {end phrase}. No other words should follow this phrase", or "The entire output should be wrapped in JSON format". This benchmark evaluates adherence to instructions rather than the content of the response.

- Big Bench Hard (BBH; Suzgun et al., 2022) – a set of 6,511 items grouped into 23 tasks from the Big Bench benchmark, focusing on the most challenging problems for LLMs. These tasks include multistep arithmetic, algorithmic reasoning (e.g., Boolean expressions, SVG shapes), language understanding (e.g., sarcasm detection, name disambiguation), and world knowledge.

- MATH lvl 5 (Hendrycks et al., 2021) – subset of the MATH benchmark consisting of 12,500 items (7,500 training and 5,000 test items) based on high-school-level competition problems gathered from various sources. Items are consistently formatted using LaTeX for equations and Asymptote for figures. The benchmark is categorized into five levels of difficulty and seven content areas (Prealgebra, Algebra, Number Theory, Counting and Probability, Geometry, Intermediate Algebra, and Precalculus). The leaderboard used here includes only items from difficulty level 5, giving this subset its name.

- Graduate-Level Google-Proof Q&A Benchmark (GPQA; Rein et al., 2023) – a set of 448 highly challenging knowledge questions crafted by PhD-level domain experts in fields such as biology, physics, and chemistry. The questions are designed to be difficult for laypersons (even with access to Google) but relatively easy for experts.

- Multistep Soft Reasoning (MuSR; Sprague et al., 2023) – a benchmark consisting of algorithmically generated complex problems, presented in narratives approximately 1,000 words in length that leverage long-range context parsing. The problems are divided into three categories: murder mysteries (250), object placement questions (256), and team allocation optimizations (250).

- Massive Multitask Language Understanding – Professional (MMLU-PRO; Wang et al., 2024) – an expert-refined version of the MMLU benchmark, consisting of 12,032 multiple-choice items (with 10 response alternatives per item) across 14 areas: math, physics, chemistry, law, engineering, economics, health, psychology, business, biology, computer science, philosophy, and miscellaneous.

Importantly, four of six benchmarks in Study 2 include an anti-guessing correction in the measure of LLM performance, as the majority of the questions are in an MC format. This correction is based on setting the baseline probability of a randomly selected answer as $P(U_i = 1) = 1/O_i$, where $U_i$ is the score on item $i$, and $O_i$ is the number of response options for item $i$. This adjustment leads to two substudies in Study 2: Study 2a and Study 2b, which focus on investigating the structure of the second benchmark using unnormalized (raw) and normalized (corrected) scores for the four benchmarks, respectively.

## 5. Results

### 5.1 Study 1 – Analysis of the Hugging Face Leaderboard v. 1

The initial unidimensional model calibrated on the older leaderboard dataset, exhibited relatively poor model fit under Maximum Likelihood Robust estimator (Yuan & Bentler, 2000). Specifically, SRMR = 0.054, robust RMSEA = 0.304 (90% CI for RMSEA = [0.294, 0.314]), CFI = 0.901, TLI = 0.836 (the baseline model scaled $\chi^2$-statistic = 10,613.699, Yuan-Bentler correction factor = 3.010, degrees of freedom for $\chi^2$ = 15, p-value < 0.001; the tested model scaled $\chi^2$-statistic = 1,332.422, Yuan-Bentler correction factor = 2.372, degrees of freedom for $\chi^2$ = 9, p-value < 0.001). Despite the poor model fit, all standardized factor loadings were exceptionally high and statistically significant (the lowest z-value = 47.101). The parameter estimates are presented in Table 1.

**Table 1**

*Parameter estimates from the initial model*

| Benchmark | Factor loading | | | Intercept | | | Residual variance | | |
|---|---|---|---|---|---|---|---|---|---|
| | UnStd | | Std | UnStd | | Std | UnStd | | Std |
| | Est. | S.E. | | Est. | S.E. | | Est. | S.E. | |
| ARC | 13.928 | 0.146 | 0.997 | 54.003 | 0.227 | 3.867 | 1.017 | 0.735 | 0.005 |
| HellaSwag | 15.239 | 0.258 | 0.932 | 73.862 | 0.266 | 4.517 | 35.186 | 1.247 | 0.132 |

| MMLU | 14.89 | 0.139 | 0.918 | 52.381 | 0.263 | 3.231 | 41.094 | 1.31 | 0.156 |
| TruthfulQA | 6.451 | 0.137 | 0.652 | 49.589 | 0.161 | 5.011 | 56.320 | 1.326 | 0.575 |
| Winograde | 9.23 | 0.118 | 0.957 | 72.298 | 0.157 | 7.499 | 7.759 | 0.700 | 0.083 |
| GSM8K | 18.857 | 0.235 | 0.740 | 29.588 | 0.414 | 1.161 | 293.761 | 5.855 | 0.452 |

TruthfulQA appears to exhibit a substantial amount of factor-irrelevant variance, suggesting that while most of the other benchmarks measure a similar property (possibly excluding GSM8K), TruthfulQA reflects a distinct aspect of LLM performance. This observation aligns with its purpose: while other benchmarks primarily capture the "cognitive" aspects of LLMs, TruthfulQA focuses on measuring LLM robustness against various biases.

The high factor loadings on the general factor are consistent with previous research attempting to apply an EFA model to similar data. For instance, Ilić (2023) provided strong evidence supporting a single general factor of intelligence across various LLMs and leaderboards, using SRMR as the model fit statistic. Similarly, Perlitz et al. (2024) demonstrated a relatively high degree of agreement between different benchmarks, although some benchmarks were less correlated with the majority than others. However, our analysis indicates that despite strong and positive factor loadings and low SRMR, the unidimensional model generally exhibits poor fit. This finding suggests that the multivariate distribution of LLM performance data is more complex than initially expected.

To improve model fit, we employed a "greedy algorithm" strategy. This involved (1) analyzing model modification indices, (2) adding the residual covariance parameter that promised the greatest improvement in model fit, (3) recalibrating the model, and (4) repeating the process. On the third iteration, the model suggested adding a residual covariance between the WinoGrande and ARC benchmarks. However, this addition resulted in negative residual variance estimates for ARC, indicating poor model convergence. Therefore, Table 2 reports only the results from the second iteration.

The revised model showed improved fit across most indices (SRMR = 0.041, robust RMSEA = 0.209 (90% CI for RMSEA = [0.197, 0.221]), CFI = 0.964, TLI = 0.922; the tested model scaled $\chi^2$-statistic = 554.775, Yuan-Bentler correction factor = 2.106, degrees of freedom for $\chi^2$ = 7, p-value < 0.001).

**Table 2**

*Residual correlations in the revised model in the order of addition*

| Benchmark | Factor loading | | Intercept | | Residual variance | |
|---|---|---|---|---|---|---|
| | UnStd | Std | UnStd | Std | UnStd | Std |

| | Est. | S.E. | | Est. | S.E. | | Est. | S.E. | |
|---|---|---|---|---|---|---|---|---|---|
| ARC | 13.916 | 0.146 | 0.997 | 54.003 | 0.227 | 3.867 | 1.349 | 0.694 | 0.007 |
| HellaSwag | 15.277 | 0.257 | 0.934 | 73.862 | 0.266 | 4.517 | 34.033 | 1.371 | 0.127 |
| MMLU | 14.872 | 0.14 | 0.917 | 52.381 | 0.263 | 3.231 | 41.634 | 1.323 | 0.158 |
| TruthfulQA | 6.489 | 0.133 | 0.656 | 49.589 | 0.161 | 5.011 | 55.834 | 1.258 | 0.570 |
| Winograde | 9.239 | 0.117 | 0.958 | 72.298 | 0.157 | 7.499 | 7.582 | 0.665 | 0.082 |
| GSM8K | 18.786 | 0.232 | 0.737 | 29.588 | 0.414 | 1.161 | 296.438 | 5.752 | 0.457 |

| Residual covariances | | | | |
|---|---|---|---|---|
| Benchmarks | UnStd | | Std | |
| | Est. | S.E. | | |
| HellaSwag ~ TruthfulQA | -24.601 | 0.799 | -0.564 | |
| MMLU ~ GSM8K | 44.408 | 1.845 | 0.400 | |

Adding these residual correlations resulted in the improvement in AIC at each step (from 157258.794 to 155909.493, to 155270.746), indicating a better model fit with each iteration. Lower AIC values signify improved relative model fit, even after accounting for the penalty for additional model parameters. The decision to use AIC instead of information criteria that incorporate sample size (e.g., Bayesian Information Criterion; Schwarz, 1978) stems from the tendency of such criteria to oversimplify the data-generating model (Evans, 2019), particularly in the context of IRT (Robitzsch, 2022). Importantly, none of the factor loadings on the general factor became insignificant or showed a substantial decrease in standardized estimates after the inclusion of these parameters.

Regarding the interpretation of the added residual covariances, all are meaningful. The negative correlation between HellaSwag and TruthfulQA suggests that vaguely defined common sense (HellaSwag) may conflict with factual accuracy (TruthfulQA). This is consistent with TruthfulQA's design, which aims to detect such contradictions. As a result, a model performing well on common sense questions might perform worse on TruthfulQA. Additionally, the positive correlation between MMLU and GSM8K can be attributed to the presence of STEM-related items in MMLU. These items share substantial common variance with the mathematics items in GSM8K, as both measure the same mathematical ability of LLMs.

Overall, the unidimensional model supports the presence of a single general factor across all benchmarks. This implies that the benchmarks measure, to some extent, the same underlying ability. Specifying a multidimensional model to approximate the variance-covariance matrix with additional factors would be futile for two reasons.

First, the specification of latent factors should be grounded in a substantial theoretical hypothesis about the latent variables underlying the observed data. Without a robust theoretical framework, such specification is not feasible. Second, given the high factor loadings on the latent factor in the unidimensional model, the correlation between factors in a multidimensional model would likely approach unity, causing convergence issues. This finding is consistent with Ilić (2023).

However, other studies suggest that LLM performance can be described using three factors: reasoning, comprehension, and core language modeling (Burnell et al., 2023), or by several principal components (Ruan et al., 2024). Our results may also partially support these findings. Specifically, our results imply that while a general factor is present in the data, residual dependencies related to task content also exist.

### 5.2.1 Study 2a – Analysis of the Hugging Face Leaderboard v. 2 (Raw Data)

The initial unidimensional model calibrated with Maximum Likelihood Robust estimator on Hugging Face Leaderboard dataset exhibited a somewhat better model fit than the model the Leaderboard v.1. Particularly, SRMR = 0.038, robust RMSEA = 0.134 (90% CI for RMSEA = [0.118, 0.151]), CFI = 0.965, TLI = 0.941 (the baseline model scaled $\chi^2$-statistic = 2599.826, Yuan-Bentler scaling factor = 2.397, degrees of freedom for $\chi^2$ = 15, p-value < 0.001; the tested model scaled $\chi^2$-statistic = 174.291, Yuan-Bentler scaling factor = 1.323, degrees of freedom for $\chi^2$ = 9, p-value < 0.001). The parameter estimates are presented in Table 3.

**Table 3**

*Parameter estimates from the initial model*

| Benchmark | Factor loading | | | Intercept | | | Residual variance | | |
|---|---|---|---|---|---|---|---|---|---|
| | UnStd | | Std | UnStd | | Std | UnStd | | Std |
| | Est. | S.E. | | Est. | S.E. | | Est. | S.E. | |
| IFEval | 0.663 | 0.029 | 0.587 | -0.399 | 0.031 | -0.353 | 0.838 | 0.166 | 0.656 |
| BBH | 0.432 | 0.008 | 0.981 | -0.131 | 0.012 | -0.297 | 0.007 | 0.002 | 0.037 |
| Math | 1.542 | 0.088 | 0.513 | -4.032 | 0.082 | -1.340 | 6.671 | 0.446 | 0.737 |
| GPQA | 0.141 | 0.004 | 0.845 | -0.894 | 0.005 | -5.345 | 0.008 | 0.000 | 0.287 |
| MuSR | 0.103 | 0.004 | 0.604 | -0.398 | 0.005 | -2.329 | 0.019 | 0.001 | 0.636 |
| MMLU-PRO | 0.595 | 0.011 | 0.955 | -0.892 | 0.017 | -1.431 | 0.034 | 0.002 | 0.088 |

All factor loadings were statistically significant, with the lowest z-value being 17.439.

Following the greedy model improvement strategy used in Study 1, we investigated model modification indices. After five iterations, an acceptable model fit was achieved. The final model showed SRMR = 0.007, RMSEA = 0.057 (90% CI for RMSEA = [0.036, 0.081]), CFI = 0.997, TLI = 0.989 (scaled $\chi^2$-statistic = 18.411, Yuan-Bentler scaling factor = 1.156, degrees of freedom for $\chi^2$ = 4, p-value = 0.001). The parameter estimates are presented in Table 4.

**Table 4**

*Residual correlations in the revised model in the order of addition*

| Benchmark | Factor loading | | | Intercept | | | Residual variance | | |
|---|---|---|---|---|---|---|---|---|---|
| | UnStd | | Std | UnStd | | Std | UnStd | | Std |
| | Est. | S.E. | | Est. | S.E. | | Est. | S.E. | |
| IFEval | 0.671 | 0.029 | 0.593 | -0.399 | 0.031 | -0.353 | 0.828 | 0.166 | 0.648 |
| BBH | 0.433 | 0.008 | 0.983 | -0.131 | 0.012 | -0.297 | 0.006 | 0.002 | 0.033 |
| Math | 1.491 | 0.087 | 0.496 | -4.032 | 0.082 | -1.342 | 6.801 | 0.444 | 0.754 |
| GPQA | 0.141 | 0.004 | 0.845 | -0.894 | 0.005 | -5.345 | 0.008 | 0.000 | 0.287 |
| MuSR | 0.102 | 0.004 | 0.599 | -0.398 | 0.005 | -2.329 | 0.019 | 0.001 | 0.641 |
| MMLU-PRO | 0.594 | 0.011 | 0.953 | -0.892 | 0.017 | -1.431 | 0.036 | 0.002 | 0.093 |
| Residual covariances | | | | | | | | | |
| Benchmarks | | | UnStd | | | | Std | | |
| | | | Est. | | S.E. | | | | |
| GPQA ~ MuSR | | | 0.003 | | 0.000 | | 0.219 | | |
| Math ~ MMLU-PRO | | | 0.103 | | 0.017 | | 0.209 | | |
| IFEval ~ GPQA | | | -0.017 | | 0.002 | | -0.207 | | |
| IFEval ~ MuSR | | | -0.020 | | 0.003 | | -0.162 | | |
| IFEval ~ Math | | | 0.291 | | 0.113 | | 0.123 | | |

Further addition of residual covariances was deemed unnecessary, as acceptable fit was achieved according to all criteria. All factor loadings remained statistically significant, with the lowest z-value being 17.068. Each added residual parameter improved the relative model fit, as reflected in the AIC values, which decreased from 7274.522 to 7210.461, to 7162.537, to 7127.051, to 7095.328, and finally to 7075.246. However, the reliability of the factor score estimates was 0.579, which is relatively low. This indicates that the unnormalized data provides relatively imprecise (in a practical sense) estimates of LLM ability.

The interpretation of the residual correlations is challenging. For example, the positive correlation between GPQA and MuSR is somewhat puzzling. While GPQA assesses expert-level knowledge across various disciplines, MuSR tests the ability to work with long context windows. The most plausible explanation for this correlation is that expert-level knowledge inherently requires the ability to retain and process a large amount of contextual information, even if it is not explicitly present in the question.

Similar to the findings in Study 1, the mathematical benchmark correlates with the results of MMLU(-PRO). Again, this can be attributed to the fact that MMLU(-PRO) contains a substantial number of STEM-related questions, which require mathematical abilities from LLMs.

The correlations of IFEval with three other benchmarks are more difficult to interpret. Notably, IFEval shows negative correlations with GPQA and MuSR, suggesting that models tend to excel at either IFEval or the other two benchmarks, but not both. This likely reflects the distinct nature of these tasks: IFEval requires models to follow explicit instructions that are largely independent of the content of the response. In contrast, GPQA and MuSR demand an understanding of the content itself, regardless of its formal or superficial characteristics. Additionally, GPQA leverages contextual knowledge beyond the information explicitly provided in the question. These differences seem to make the characteristics required for IFEval and the other two benchmarks inherently incompatible to some degree.

The positive correlation between IFEval and the math benchmark, however, can be explained by the shared requirement for models to handle technical language such as LaTeX effectively. This suggests that the ability to work with equations represents a somewhat distinct skill in LLMs.

Overall, these results again indicate the presence of a single factor of general ability across all benchmarks, supporting findings from Ilić (2023).

## 5.2.2 Study 2b – Analysis of the Hugging Face Leaderboard v. 2 (Normalized Data)

The initial unidimensional model, calibrated using the Maximum Likelihood Robust estimator on the Hugging Face Leaderboard dataset, exhibited somewhat better model fit compared to the initial models in Studies 1 and 2a. Specifically, SRMR = 0.028, robust RMSEA = 0.092 (90% CI for RMSEA = [0.073, 0.112]), CFI = 0.979, TLI = 0.965 (the baseline model scaled $\chi^2$-statistic = 807.201, Yuan-Bentler scaling factor = 6.179, degrees of freedom for $\chi^2$ = 15, p-value < 0.001; the tested model scaled $\chi^2$-statistic = 96.850, Yuan-Bentler scaling factor = 1.220, degrees of freedom for $\chi^2$ = 9, p-value < 0.001). The parameter estimates are presented in the table 5.

**Table 5**

*Parameter estimates from the initial model*

| Benchmark | Factor loading | | | Intercept | | | Residual variance | | |
|---|---|---|---|---|---|---|---|---|---|
| | UnStd | | Std | UnStd | | Std | UnStd | | Std |
| | Est. | S.E. | | Est. | S.E. | | Est. | S.E. | |
| IFEval | 0.668 | 0.032 | 0.591 | -0.399 | 0.031 | -0.353 | 0.832 | 0.163 | 0.651 |
| BBH | 0.980 | 0.022 | 0.970 | -1.318 | 0.027 | -1.304 | 0.061 | 0.015 | 0.060 |
| Math | 1.706 | 0.092 | 0.567 | -4.032 | 0.082 | -1.340 | 6.140 | 0.436 | 0.678 |
| GPQA | 1.609 | 0.083 | 0.652 | -3.735 | 0.067 | -1.514 | 3.498 | 0.268 | 0.575 |
| MuSR | 0.450 | 0.020 | 0.547 | -2.497 | 0.022 | -3.040 | 0.472 | 0.065 | 0.700 |
| MMLU-PRO | 1.150 | 0.029 | 0.955 | -1.529 | 0.033 | -1.269 | 0.128 | 0.046 | 0.088 |

All factor loadings were statistically significant, with the lowest z-value was 18.555.

Again, following the greedy model improvement strategy used in Study 1 and Study 2a, we investigated model modification indices. After four iterations, good model fit was achieved. The final model fit indices were SRMR = 0.016, RMSEA = 0.052 (90% CI for RMSEA = [0.028, 0.079]), CFI = 0.996, TLI = 0.989 (scaled $\chi^2$-statistic = 25.262, Yuan-Bentler scaling factor = 1.001, degrees of freedom for $\chi^2$ = 5, p-value < 0.001). The parameter estimates are presented in Table 6.

**Table 6**

*Residual correlations in the revised model in the order of addition*

| Benchmark | Factor loading | | | Intercept | | | Residual variance | | |
|---|---|---|---|---|---|---|---|---|---|
| | UnStd | | Std | UnStd | | Std | UnStd | | Std |
| | Est. | S.E. | | Est. | S.E. | | Est. | S.E. | |
| IFEval | 0.673 | 0.033 | 0.595 | -0.399 | 0.031 | -0.353 | 0.826 | 0.162 | 0.646 |
| BBH | 0.980 | 0.023 | 0.970 | -1.318 | 0.027 | -1.304 | 0.060 | 0.017 | 0.059 |
| Math | 1.710 | 0.093 | 0.568 | -4.032 | 0.082 | -1.340 | 6.125 | 0.439 | 0.677 |
| GPQA | 1.564 | 0.080 | 0.634 | -3.735 | 0.067 | -1.514 | 3.641 | 0.275 | 0.598 |
| MuSR | 0.427 | 0.024 | 0.521 | -2.497 | 0.022 | -3.042 | 0.491 | 0.066 | 0.729 |
| MMLU-PRO | 1.149 | 0.029 | 0.954 | -1.529 | 0.033 | -1.269 | 0.131 | 0.042 | 0.090 |
| Residual covariances | | | | | | | | | |
| Benchmarks | | | UnStd | | | | | Std | |

|  | Est. | S.E. |  |
|---|---|---|---|
| BBH ~ MuSR | 0.039 | 0.015 | 0.225 |
| GPQA ~ MMLU-PRO | 0.125 | 0.041 | 0.181 |
| IFEval ~ MuSR | -0.068 | 0.014 | -0.107 |
| GPQA ~ MuSR | 0.130 | 0.038 | 0.097 |

Further addition of residual covariances was deemed unnecessary, as acceptable fit according to all criteria had been achieved. All factor loadings remained statistically significant, with the lowest z-value being 18.056. Each added residual parameter improved the relative model fit, as indicated by the AIC values, which decreased from 23,980.931 to 23,943.2, to 23,923.77, to 23,908.108, and finally to 23,896.045. The reliability of the factor scores was 0.789, which is moderate and higher than that observed with the raw data in Study 2a. This suggests that the anti-guessing corrections and normalizations applied to the benchmark scores effectively suppressed random noise, improving the signal-to-noise ratio in the data. However, for high-stakes decisions, greater reliability is needed to achieve lower standard errors in ability estimation.

Interestingly, the structure of residual correlations obtained from the normalized data differs from that obtained from the raw data. In this case, MuSR emerges as the benchmark exhibiting the most residual correlations, indicating that it shares the greatest amount of common variance with other variables. One possible interpretation of the positive correlation between BBH and MMuSR is that it likely reflects advanced logical, or multi-step reasoning demands that both benchmarks share. In other words, BBH tasks often require the same kinds of multi-step reasoning skill that MuSR explicitly targets.

The positive correlation between GPQA and MMLU-PRO exists probably because both benchmarks revolve around knowledge retrieval and application. GPQA covers broad or general knowledge, while MMLU-PRO focuses on deep, specialized knowledge. Even beyond the overall factor (which captures general model ability), these two have overlapping demands: a model that is extra-strong (or extra-weak) at one tends to be similarly extra-strong (or -weak) at the other, reflecting a shared reliance on accurate knowledge retrieval and domain application.

The negative correlation between IFEval and MuSR is interesting. We hypothetize that it occurs due to two possible reasons: (1) a trade-off in how the model is fine-tuned: some models are heavily optimized for strict compliance and short, direct answers, while others are optimized more for open-ended reasoning; and (2) differences in how these tasks are structured.

Specifically, IFEval may reward concise adherence to instructions, whereas MuSR may demand a more exploratory chain-of-thought.

Finally, the positive correlation between GPQA and MuSR can be explained by the following reasons. GPQA questions sometimes require multiple reasoning steps (e.g., multi-hop QA). Thus, there's an extra positive link: doing well on multi-step reasoning tasks beyond general LLM skill also translates into doing well on open-ended QA that calls on those same reasoning processes. Models that excel at carefully chaining facts and logic tend to show an extra edge in general QA tasks that demand multi-hop reasoning.

Overall, the differences between the Raw and Normalized data residual correlations implies that random noises can disguise the "meaningful" covariance pattern. Also, the residual covariances in the normalized data are more difficult to interpret than those in the raw data in Study 2a. This could be due to two factors: (1) the vague definitions of the abilities that the benchmarks measure, which hinder clear and straightforward interpretation of LLMs' world models, and/or (2) the possibility that LLMs' world models differ significantly from those of humans.

Notably, the pattern of standardized factor loadings and standardized residual variances among the observed variables remains stable between the normalized and raw data. The variance in between-LLM performance for the BBH and MMLU-PRO benchmarks is almost entirely explained by the latent ability factor. This could be attributed to the fact that these benchmarks inherently reflect LLM performance across a diverse set of tasks, analogous to the general factor of intelligence (IQ g-factor), which can be roughly defined as the ability to execute a wide range of diverse tasks. Consequently, these measures are the closest representations of this g-factor.

In contrast, MuSR, Math, and IFEval exhibit the largest shares of unexplained variance (and, by definition, the lowest factor loadings), as the abilities required to perform well on these benchmarks are more specialized and less generalizable.

## 5.3. Study 3 – The Comparison of Estimated Factor Scores and the Average Scores on Benchmarks

To compare LLM rankings, we use scatterplots with a green trend line predicted by B-splines with three internal knots positioned at the $25^{th}$, $50^{th}$, and $75^{th}$ percentiles of the x-axis variable. For plots where the x-axis represents the number of model parameters (in billions), we include an additional $4^{th}$ knot at 70 billion parameters. This adjustment accounts for the gamma distribution of model parameters, which has a natural lower bound of 0 and a long right tail. Consequently, the leaderboard contains many relatively small models, with the number of models decreasing as parametric complexity increases. Moreover, the distribution of models by parameter count shows a mode near the 70 billion parameter mark, justifying the placement of the fourth knot in these plots.

Around the green trend line, the blue area represents the prediction interval of *±2 standard errors of B-spline prediction accuracy*. In some plots, red whiskers around each dot indicate *±1 standard error of factor scores* (as described in Equation 14).

**Figure 1**

*Comparison of the Factor Scores and Benchmark Averages from the first version of the Leaderboard*

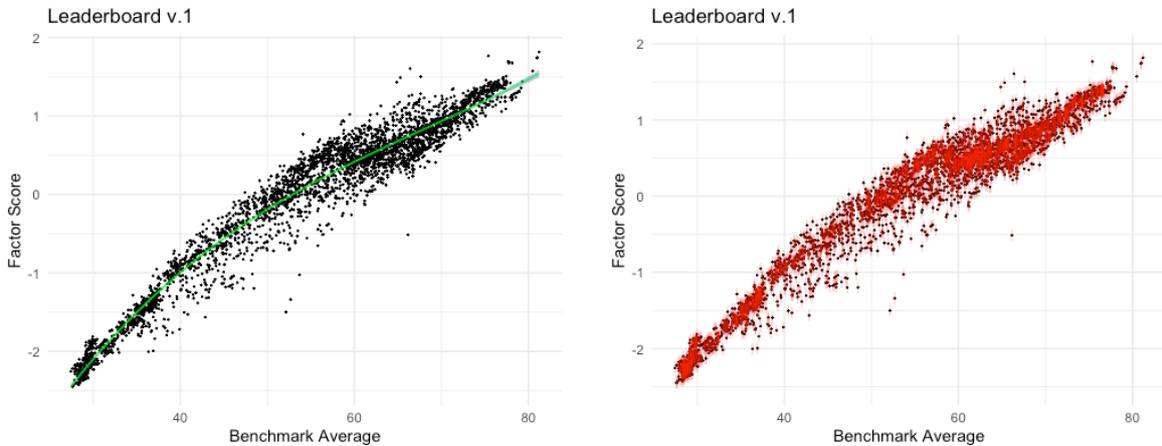

A. With the S.E.s of spline prediction          B. With the S.E.s of factor scores

The comparison between the native LLM ranking from the Leaderboard (Benchmark Average) and the ranking of LLMs derived from the FA model reveals a slight inverted U-shaped dependency. This pattern indicates that as LLM performance in Benchmark Averages increases, the rate of progression in Factor Scores decreases, particularly at higher ability levels.

This finding suggests that the slowdown in LLM scaling (i.e., the improvement in LLM performance; Hu & Tong, 2024) is more evident in the ability levels measured by Factor Scores. In contrast, Benchmark Averages tend to obscure – or mask the severity of – this trend, potentially underrepresenting the diminishing returns observed in LLM scaling efforts.

Figure 1B is arguably more important than Figure 1A, as the standard errors of factor scores reveal that many models do not differ statistically significantly. Consequently, ranking LLMs solely based on the Benchmark Average may suggest that some models outperform others, even though their abilities are not statistically significantly different.

**Figure 2**

*Comparison of the LLM performance and the number of LLMs parameters in billions from the first version of the Leaderboard*

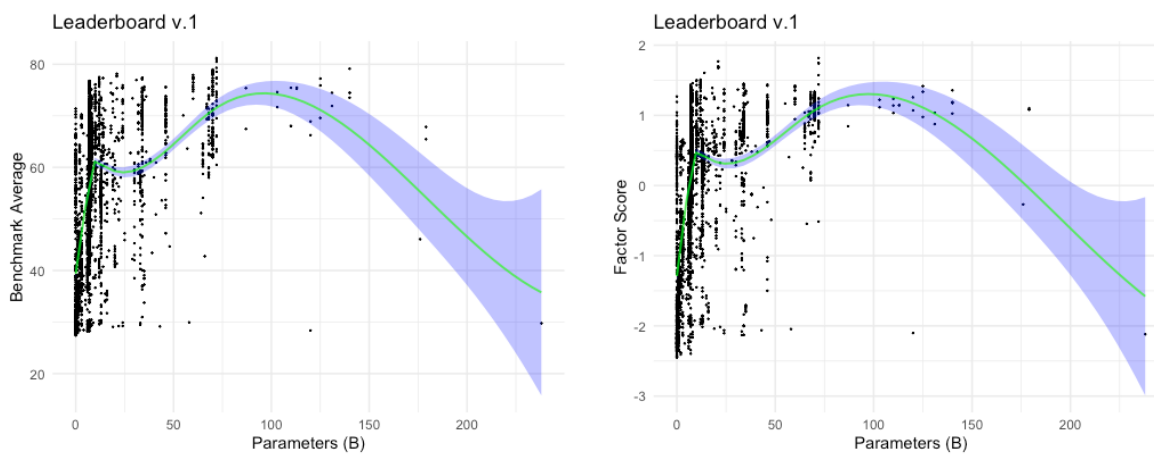

A. For the Benchmark Average          B. For the factor scores

The comparison of Figures 2A and 2B reveals that the improvement in LLM with the increase in further parametric complexity performance slows down. Due to outliers in the data (i.e., massive models with unexpectedly low performance), the trend line declines as model size increases, accompanied by increasingly large standard errors. However, the overall trend remains observable.

Overall, these results suggest that the FA model introduces a meaningful correction to the model rankings used in the Leaderboard by "filtering out" individual noise from specific benchmarks and ranking LLMs based on the true common variance across all benchmarks. While the top-performing models in the Leaderboard remain the same, the rank order of the "average" LLMs can change significantly.

However, the FA model for Leaderboard v.1 failed to converge under the improvement modifications, and the reported factor scores are from a model that generally fits the data poorly. Because of this limitation, we proceed to the analysis of Leaderboard v.2.

**Figure 3**

*Comparison of the Factor Scores on the raw data and Benchmark Averages from the second version of the Leaderboard*

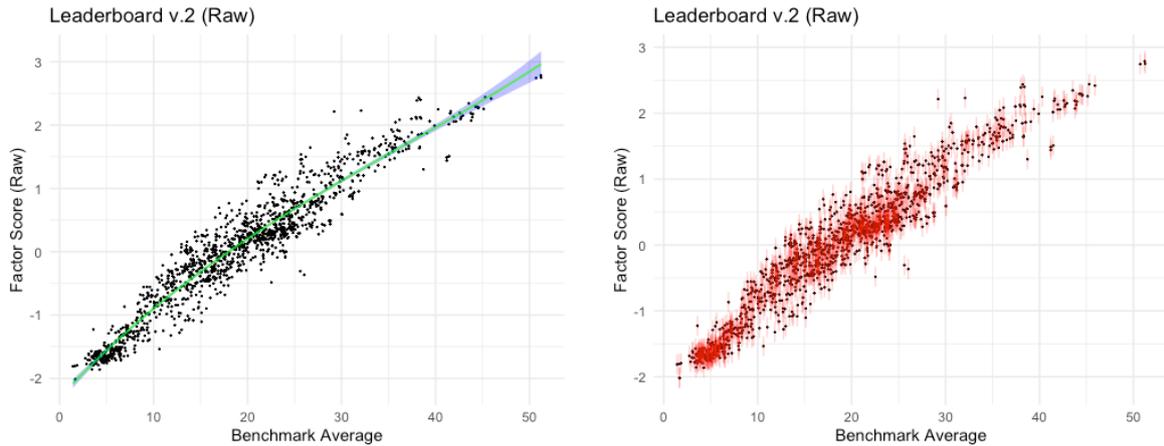

A. With the S.E.s of spline prediction      B. With the S.E.s of factor scores

Figure 3 shows that on the raw data, the correction introduced by factor scores in LLM rankings is generally minimal. While some models are under- or overestimated in terms of their ability when comparing factor scores to benchmark averages (particularly among average models), the trend remains fairly linear.

**Figure 4**

*Comparison of the Factor Scores on the normalized data and Benchmark Averages from the second version of the Leaderboard*

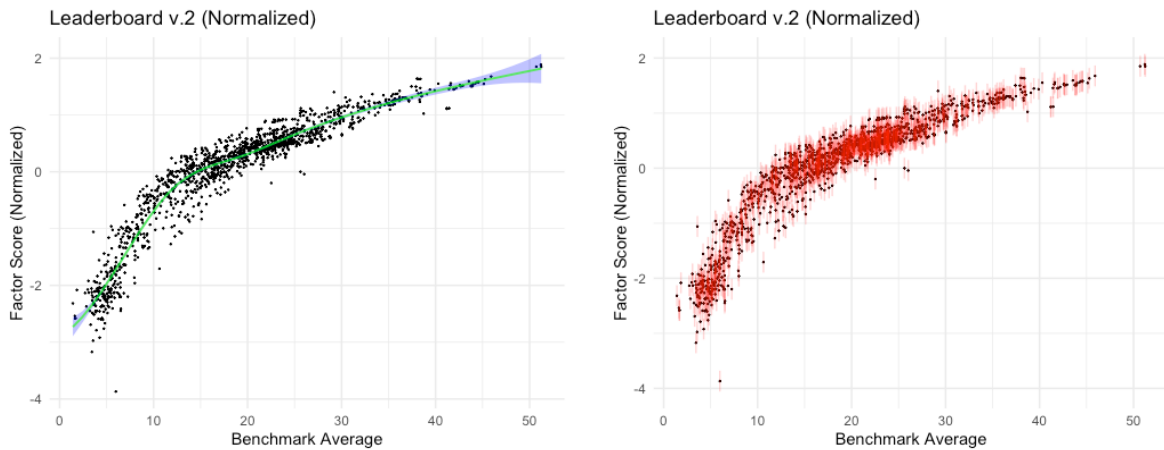

A. With the S.E.s of spline prediction      B. With the S.E.s of factor scores

In contrast, when comparing factor scores on the normalized data in Figure 4, an inverted U-shaped trend (similar to the results from the first Leaderboard) becomes visible. This indicates that factor scores, as a measure of LLM ability on normalized data, penalize weaker-performing models more heavily compared to medium- and high-performing models. This behavior aligns with the characteristics of IRT models such as the 3PL, which penalize weaker respondents more strongly under the assumption that their scores are primarily due to random guessing.

These findings are further supported by comparisons of factor scores derived from raw and normalized data models (see Fig. 5).

**Figure 5**

*Comparison of the Factor Scores on the raw and normalized data*

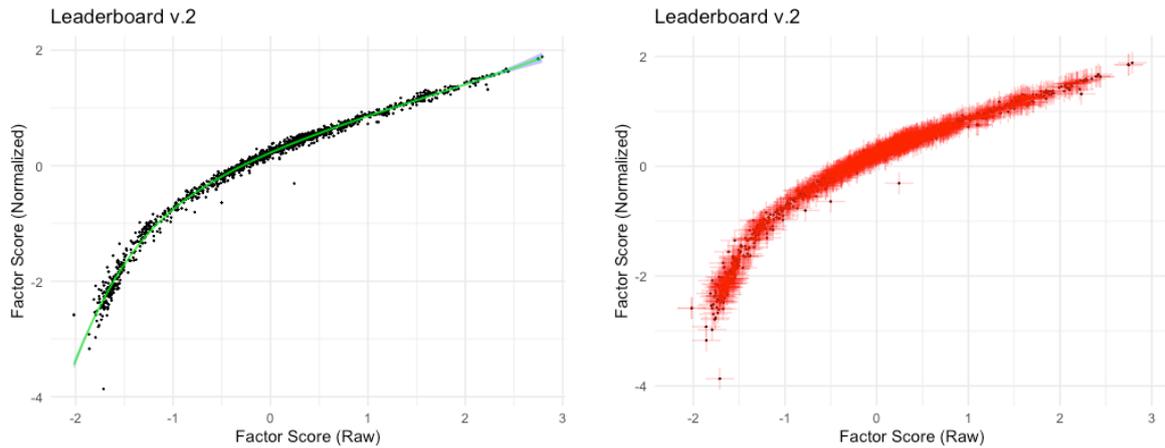

A. With the S.E.s of spline prediction          B. With the S.E.s of factor scores

**Figure 6**

*Comparison of the LLM performance and the number of LLMs parameters in billions from the second version of the Leaderboard*

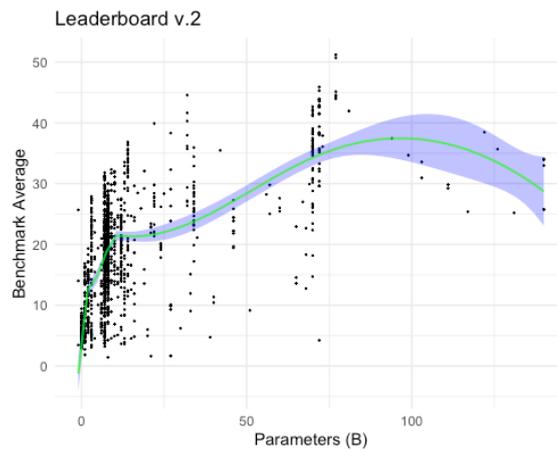

A. For the Benchmark Average

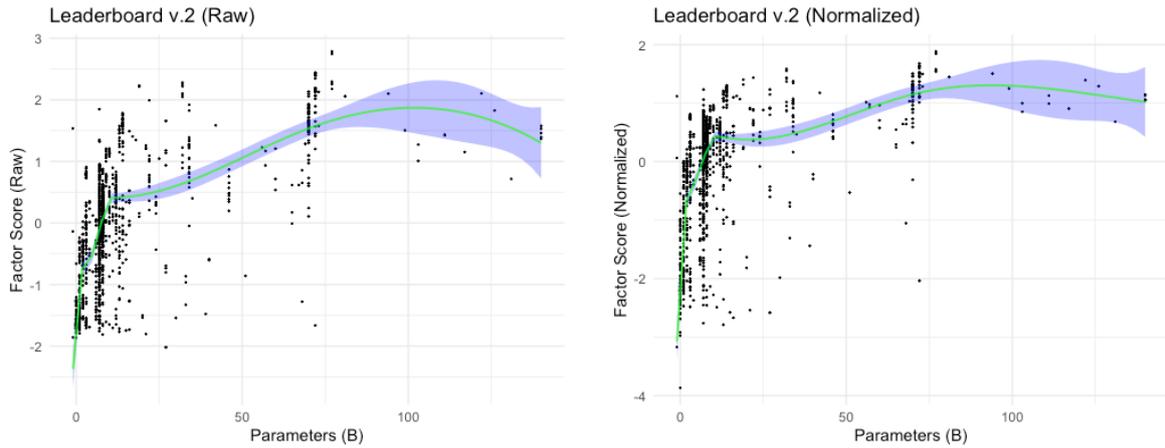

B. For the factor scores on the raw data

C. For the factor scores on the normalized data

The comparison of Figures 6A, 6B, and 6C, again, highlights the relatively recent observation that LLM scaling (Hu & Tong, 2024) has reached a performance ceiling. A clear ceiling effect indicates that further increases in the number of parameters in LLMs results in increasingly diminishing improvements in model performance. However, this trend is most apparent in the normalized factor scores and most confusing in the Benchmark Average.

**Figure 7**

*Comparison of the LLM performance and the grams of CO₂-equivalent emissions during the model training from the second version of the Leaderboard*

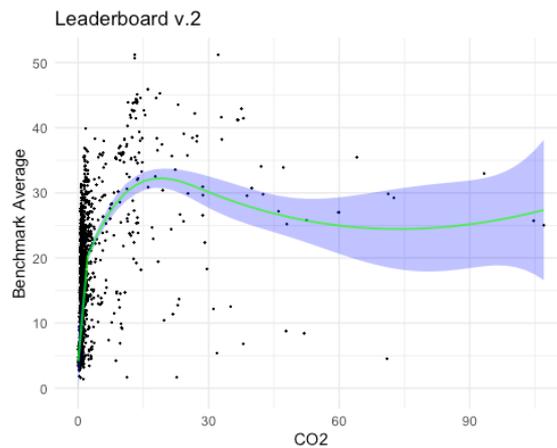

A. For the Benchmark Average

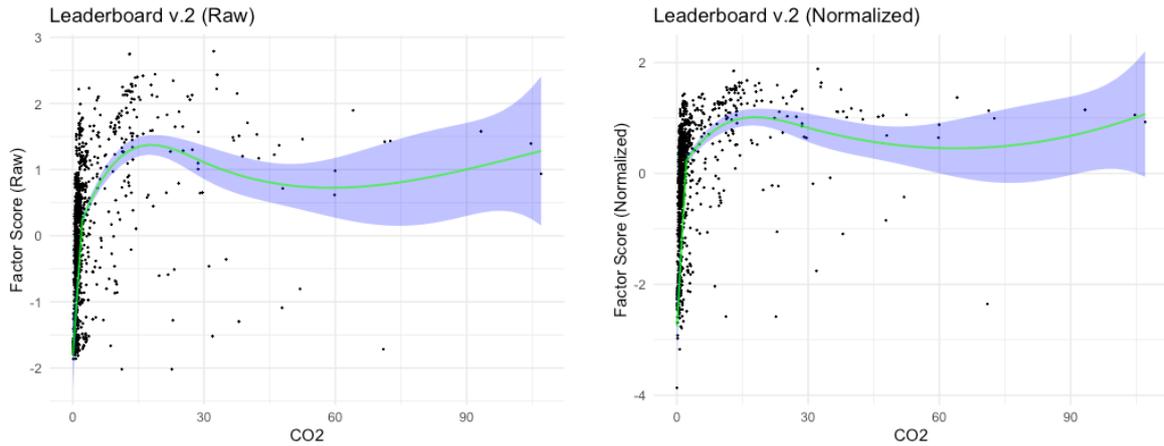

B. For the factor scores on the raw data     C. For the factor scores on the normalized data

Similarly, Figures 7A, 7B, and 7C demonstrate that increasing parametric complexity of LLMs leads to the larger environmental impact without substantive improvement in the model performance. This is expected, as the amount of $CO_2$-equivalent emissions is strongly correlated with the number of model parameters (see Fig. 8).

**Figure 8**

*Comparison of the grams of $CO_2$-equivalent emissions and the number of model parameters in billions from the second version of the Leaderboard*

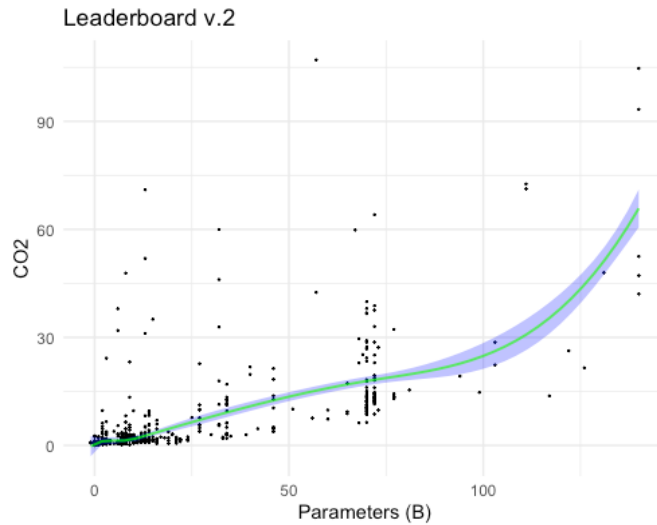

Certain practical insights can also be derived by comparing the distributions of different types of LLMs represented in the leaderboard, as done in a similar spirit to the work of Sun et al. (2024).

**Figure 9**

*Comparison of different LLMs architectures in terms of their performance*

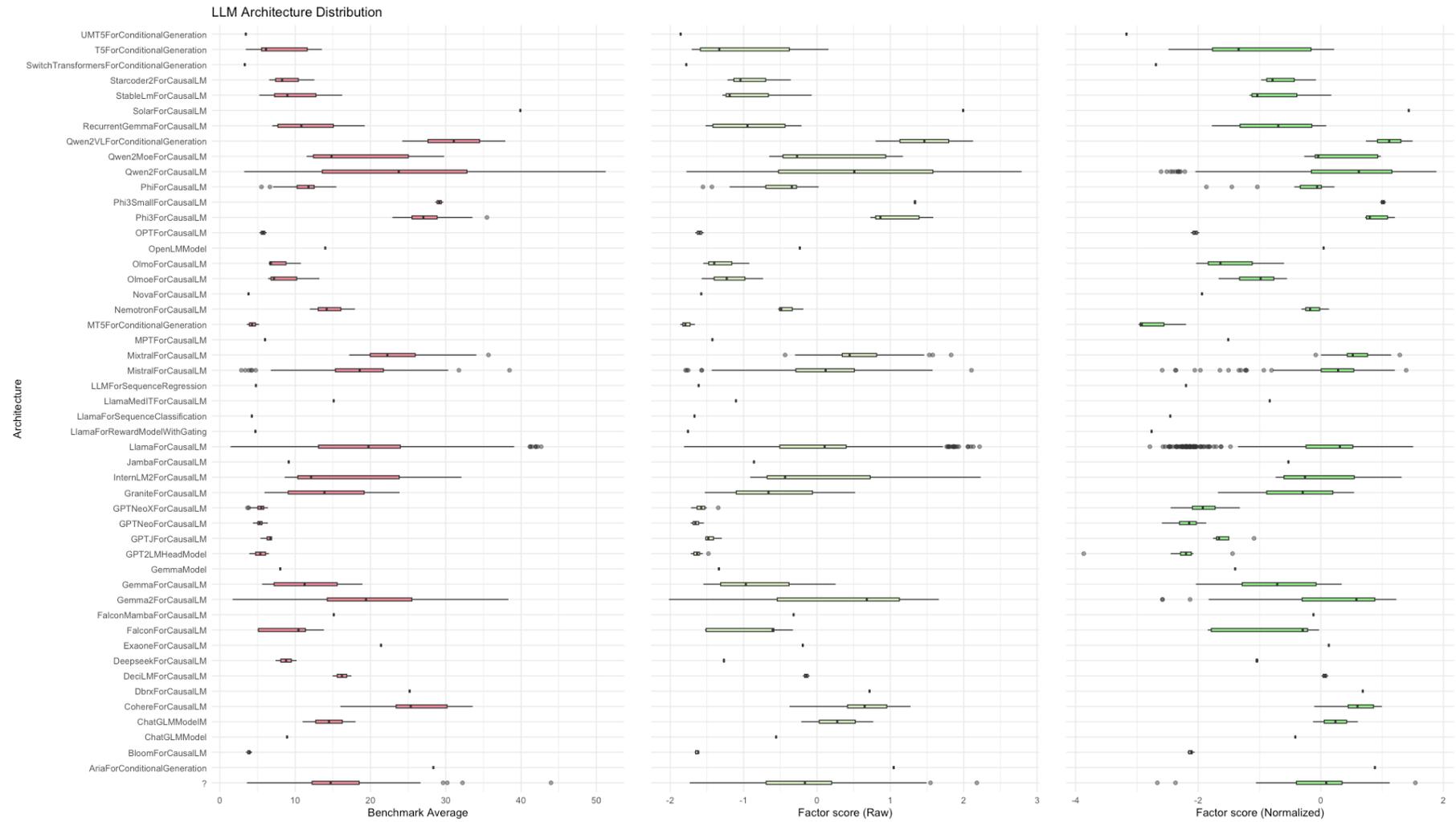

Figure 9 reveals that the variance of certain basic architectures in terms of their factor scores differs from their variance in Benchmark Averages. For instance, GraiteForCasualLM shows a performance variance similar to InternLM2ForCasualLM in factor scores, while its dispersion in Benchmark Average is smaller. By contrast, Qwen2VLForConditionalGeneration demonstrates much more homogeneous results according to factor scores compared to the Benchmark Average.

**Figure 10**

*Comparison of different LLMs types in terms of their performance*

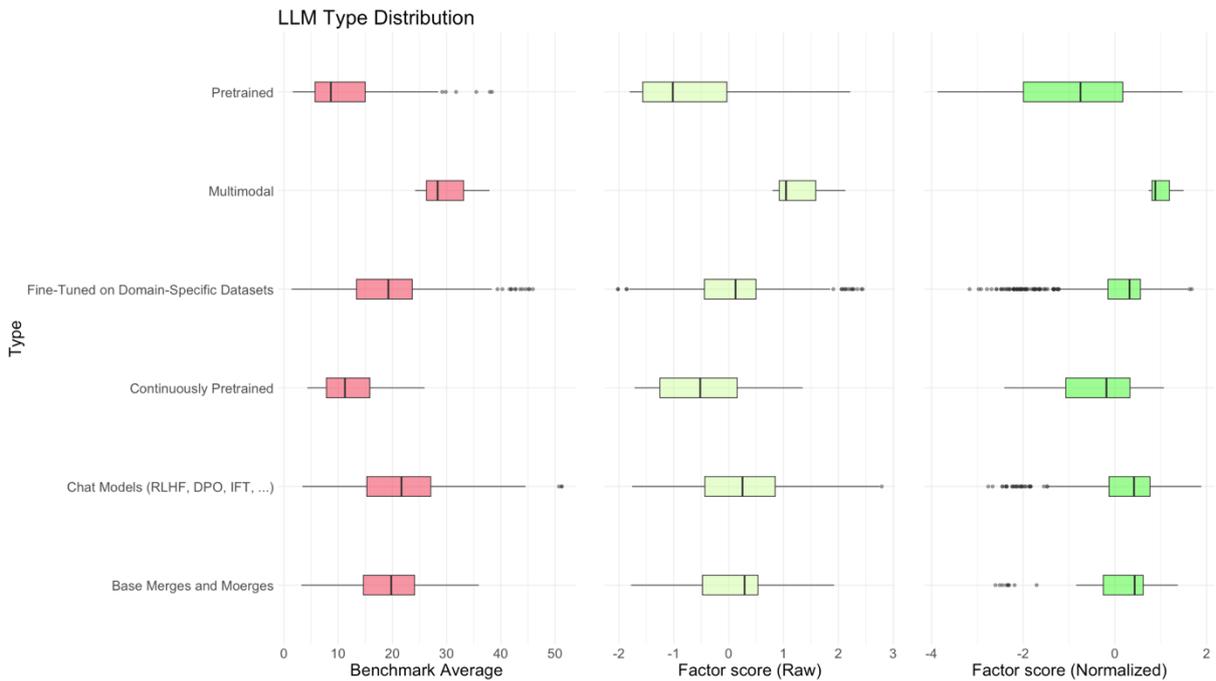

Figure 10 is also noteworthy, as it highlights certain features of model training. For example, multimodal models appear to be more consistent in their performance according to factor scores than according to Benchmark Averages. Additionally, fine-tuning on domain-specific datasets or chat-tuning tends to produce more abnormally low results in terms of factor scores than abnormally high results in terms of Benchmark Averages, presumably due to catastrophic forgetting (Parisi et al., 2019).

These observations, however, warrant further substantive discussion, which lies beyond the scope of this paper.

## 6. Conclusion

With the emergence of ChatGPT, the rise of AI tools for information processing has led to the development of myriad LLMs. Correspondingly, the question "Which LLM is the best?" has become increasingly pressing. To address this, multiple benchmarks have been developed to quantify the performance of various LLMs and rank them based on their (cognitive) capabilities.

However, while benchmarks may appear superficially similar to tests and surveys used in the social sciences, they differ significantly upon closer inspection. The most obvious difference is the number of items – benchmarks typically operate with orders of magnitude more items than surveys and tests. More importantly, benchmarks often lack a detailed description of the emergent property they aim to measure in the behavior of a (cognitive) system. In many cases, benchmark papers provide little more than a couple of sentences to describe the trait being measured. In contrast, the development of theoretical frameworks for human tests and surveys is one of the most time-intensive components of test development. Carefully defining the targeted construct, its components, the boundaries of its domain, and its relationships with other constructs are foundational steps in test design (Mislevy et al., 2003).

Nonetheless, the mathematical frameworks used to process data from social surveys and tests can be applied to leaderboards and benchmarks. Our analysis demonstrates that leaderboards and benchmarks provide parceled data (Matsunaga, 2008), where groups of homogeneous items are averaged. Traditional psychometric models can be applied to such data to extract the common variance across parcels and estimate the rank ordering of models based on the signal – the latent variables to which the observed variables can be compressed. These models form the foundation of the entire industries of testing, psychometrics, educational assessment, and psychological evaluation.

The traditional interpretation of such modeling techniques implies that latent variables are considered "causes" of the observed responses. However, relatively recent advancements in network psychometrics suggest that latent variables cannot be considered "causes" in the strict causal sense (Marsman et al., 2018). Instead, they should be viewed as convenient and compact representations of the data in a lower-dimensional space. Recent analogies between traditional psychometric models and variational autoencoders (Urban & Bauer, 2021) or collaborative filtering techniques (Bergner et al., 2022) further support these "utilitarian" approaches to data modeling strategies.

Although the clear interpretation of the estimated latent variables is challenging due to the vague definitions of what the original benchmarks aim to measure, these latent variables offer several advantages over observed scores. Among the most significant is their ability to filter out noise from the data and estimate the truly common variance across parcels, providing a more robust measure than a simple average. For a more philosophical discussion of the advantages and disadvantages of this approach, see Sijtsma et al. (2024a), Mislevy (2024), McNeish (2024), and Sijtsma et al. (2024b).

The results of our analysis suggest that the internal structure of the first version of the Leaderboard is difficult to utilize, as attempts to improve its model fit often result in non-convergent models. In contrast, the second version of the Leaderboard with raw data (without corrections for random guessing) provides an interpretable model structure. However, the second version of the leaderboard with normalized data presents a data structure that is more difficult to interpret and different from that of the raw data. This may indicate that LLM cognition is inherently poorly interpretable by humans, but it might also reflect the unclear definitions of what the benchmarks are intended to measure.

Overall, the leaderboard appears to rank LLMs based on a single characteristic – something resembling general intelligence (Ilić, 2023). Nonetheless, a clear and comprehensive definition of this factor is highly complex and requires further investigation (Gignac & Szodorai, 2024).

When comparing rankings based on the average raw score and the factor score estimated from the latent variable model, an intriguing trend emerges. The scatterplot exhibits a vaguely inverted U-shaped pattern. This suggests that the raw average slightly overestimates the weakest-performing models and/or slightly underestimates the strongest-performing models when compared to the factor score. While the recent announcement of OpenAI's o3 model indicates that the performance of LLMs still can significantly improve, now the majority of the models are significantly smaller, and their performance is expected to be improved by the architectural innovations rather than the brute force increase in the number of trained parameters.

This observation could imply that the cognitive schemas learned by LLMs differ significantly between the weakest and strongest models, with the differences between LLMs diminishing at the tails of the distribution. Alternatively, it may indicate that the scaling laws governing LLM performance, as observed at the time of data collection, were approaching their limits. Another possible explanation is the limitations of the benchmarks themselves, which may be implicitly designed to provide more precise measurements near the center of the ability distribution.

## 7. Discussion and Limitations

This study opens several new directions for further research, particularly in the application of psychometric methodologies to estimate benchmark quality. The traditional approach to benchmark development assumes that all items, regardless of their quality, should be included in benchmarks, as LLMs – as neural networks (i.e., universal approximators) – should theoretically be capable of deciphering the meaning of questions irrespective of their quality. However, over a century of psychometric research has demonstrated that some questions can be so poorly formulated that respondents with higher levels of ability have a lower probability of answering them correctly than those with lower levels of ability. Such questions can significantly contaminate ability estimates and lead to poor decision-making based on test scores. Investigating benchmarks with this perspective could improve their quality and, consequently, enhance the comparison of LLM performance in benchmarks. Improved comparisons of LLM performance could, in turn, facilitate better investigations into the factors influencing their abilities, enabling more targeted research into what enhances LLM cognitive abilities and what does not.

One of the most promising outcomes of such research could be the potential to align different leaderboards with each other. A key driver of psychometric advancements has been the need to measure *change* in longitudinal data (e.g., Little, 2024). When analyzing how a human ability or trait changes over time, issues such as ceiling effects (arising when the same set of items is administered multiple times) or training effects (when respondents remember items from previous measurements) must be addressed. To avoid these issues, the content of test items must be modified over time. However, since test items are not interchangeable, this introduces the problem of scale incomparability. Specifically, because the items differ, numerically identical test scores from different measurement occasions (based on different sets of items) represent different levels of ability.

To address this, psychometricians developed IRT and FA techniques for measuring longitudinal change by leveraging partial repetition of the same items across measurement occasions. In these approaches, the item parameters of *anchor items* are constrained to remain constant across occasions. Through these anchor items, latent variable scale comparability is established, enabling the meaningful comparison of abilities across different measurement occasions.

The field of computer science has recognized that benchmarks tend to become easier for LLMs over time. This observation motivated the introduction of the second version of the analyzed Leaderboard, which features a more difficult set of questions. Moreover, further advancements, such as the recent announcement of OpenAI's o3 model, exacerbate the susceptibility of existing leaderboards and benchmarks to ceiling effects. The natural response to this challenge has been to make benchmarks increasingly difficult, but this approach risks rendering the estimates of "general cognitive ability" incomparable and disconnected across different leaderboard versions – an issue evident in the Hugging Face Leaderboard v.1 and v.2.

However, psychometrics may offer a solution. Notably, the Big Bench benchmark is partially comparable across both versions of the leaderboard, making it a potential tool for establishing a common scale across versions. By using benchmark data on a task-by-task basis (task-by-task parcels), item parameters for tasks common to both leaderboards can be constrained to identical values. This adjustment transforms the traditional psychometric model into a longitudinal one (Wilson et al., 2012). Such a model enables the comparison of general cognitive ability estimates over time, providing a foundation for tracking continuous improvements in LLM abilities despite changes in the sets of observed variables and addressing breakthroughs and innovations. Potentially, this research direction can result in developing a measure of the general intelligence for LLMs ("AIQ"), that will be able to track the development of more capable models and adapt in terms of the task content as the older tasks are rendered "solved".

An additional benefit of utilizing benchmarks on a task-by-task basis is the reduction in the standard error of measurement for LLM ability estimates. This metric is sensitive to the number of observed variables used in the estimation process, and analyzing benchmark tasks rather than overall benchmark performance increases the number of observed variables. Furthermore, this approach facilitates a more detailed investigation of residual dependencies between tasks, enabling deeper insights into LLMs' world models.

Investigating LLMs' world models represents another, potentially even more promising direction for further research enabled by this work. A world model (Ha & Schmidhuber, 2018) is a mental representation of the space of possible inputs to a cognitive system and the relationships between them. It serves as the foundation for generating outputs – essentially, a constructed picture of the surrounding environment within the system's "mind" (with "surrounding" understood in a broad sense).

Some researchers (e.g., Pellert et al., 2024) have attempted to use human psychological tests to estimate the traits of LLMs. However, this approach has been criticized by others (see the review by Löhn et al., 2024). A central concern is that LLMs lack constructs and psychological traits in the sense that humans possess them. While we generally agree with this critique, we do not believe it renders the application of methodologies developed for studying humans entirely invalid when applied to machines.

Psychometric methods often focus less on the deliberative processes humans use to answer questions and more on dimension reduction to reflect between-individual differences in performance. Using the same methodology for the same purpose in LLM evaluation, while being mindful of these limitations, could offer an interesting and relatively straightforward approach to exploring the templates of LLMs' world models.

Traditional psychometric models aim to identify observed indicators that are "outliers" in the sense that they share less common variance with the majority of other indicators. Essentially, factor-analytical models are designed to define the boundaries of emergent, distinct domains of human behavior (commonly referred to as *constructs* in contemporary psychology). This approach dominates the study of human world models. Since factor-analytical methods excel at identifying clusters of behavioral indicators that are so highly correlated they can be considered manifestations of a single variable, they are predominantly used to delineate the boundaries of constructs. Psychologists and psychometricians routinely search for new observed indicators, determine whether they belong to distinct constructs, and study the statistical relationships between these constructs. This stream of activity aims to explicitly articulate aspects of human world models that are so intuitive to us that we rarely (or cannot) conceptualize alternative ways to represent them.

However, LLMs, as reflections and aggregations of language materials (initially) produced by humans, can construct verbal world models that differ from human representations of the world. One of the most direct ways to investigate the structure of LLM world models is by interpreting the parameters learned by the LLMs (Templeton et al., 2024). Yet, given the immense complexity of this task, an alternative approach is to apply a methodology similar to that used in psychology – attempting to infer the world model indirectly by observing responses to different inputs.

This is where the interpretation of residual correlations in factor-analytical models becomes relevant. Residual correlations can reveal peculiar dependencies between seemingly unrelated observed indicators, even after the general common variance has been accounted for. In this sense, the dependencies between normalized benchmarks in the second version of the Leaderboard are of particular interest. However, the lack of clear construct definitions in the benchmarks significantly limits the ability to interpret these dependencies. Future research could focus on designing benchmarks with clearer and more established interpretations of their scores, which would substantially advance this area of investigation.

However, this approach encounters an immediate challenge: the number of observations. All factor-analytical methods require a correlation matrix derived from a sample drawn from a population of respondents. In the context of LLMs, obtaining such a sample (let alone defining the population) is far from straightforward. LLMs are extremely complex statistical models that require vast computational resources to train and operate. As a result, benchmarking data currently serves as one of the most critical sources for this type of research.

Furthermore, another key assumption of factor-analytical models is that the factor loading structure is consistent across all respondents within the population. This assumption has been critiqued in psychological literature (De Ayala & Santiago, 2017), but it remains largely unaddressed in LLM-related research. The term *LLM* serves as an umbrella designation, encompassing a wide variety of model architectures that all share a common purpose: predicting the next token in a sequence. However, it is entirely plausible that different models – with varying base architectures, training corpora, and task-specific fine-tuning – possess fundamentally distinct world models. Consequently, the psychological meaning of the same question may differ significantly across these models.

This phenomenon is the central focus of a branch of psychometrics and psychology dedicated to investigating cross-cultural and cross-language adaptation and translation of surveys and tests (Bauer, 2023). If an item exhibits different psychometric properties (e.g., different parameter estimates) for different groups of respondents (such as males and females) with the same level of ability, this phenomenon is referred to as *Differential Item Functioning* (DIF) or a violation of *Measurement Invariance*. Items flagged for DIF undergo significant scrutiny, as they are presumed to measure not only the construct of interest but also some secondary, group-correlated noise. This noise can introduce systematic bias into ability estimates and distort comparisons of results. Consequently, the causes of such differences are carefully studied and interpreted in terms of mental or cognitive differences between groups. The ultimate goal is to purify the interpretation of final ability estimates by eliminating these contaminators.

However, such analyses require researchers to know the true groups of respondents in advance. An alternative approach is the application of mixture factor-analytical models, which assume the existence of distinct subpopulations within the sample (De Ayala & Santiago, 2017). These subpopulations differ qualitatively from one another (i.e., items exhibit different psychometric properties for each subpopulation) but only quantitatively within each group (i.e., respondents within a group differ in ability, but item parameters remain constant). These models are particularly adept at identifying qualitatively distinct subpopulations with differing construct structures.

Applying such models to LLM benchmarking data could uncover different (groups of) world models and provide insights into which architectural features lead to these variations.

This research direction holds significant potential for advancing our understanding of LLMs and their diverse cognitive architectures.

This research has several limitations. In particular, the robustness (or lack thereof) of LLM output to variations in prompting strategies and response option order is not addressed here. LLMs, especially smaller models, tend to exhibit instability when different prompting strategies are employed. Techniques such as chain-of-thought reasoning or adjustments to the number of examples provided during few-shot prompting can significantly enhance model performance on benchmarks. However, as long as the evaluation procedure is standardized and all LLMs are tested under the same conditions, this does not pose a severe issue – although it remains an interesting area for further research. Moreover, the benchmarks used in the second study differ in their default prompting strategies. Standardizing these strategies could lead to different outcomes in terms of both item parameters (benchmark structure) and LLM rankings.

Arguably, the most significant limitation of this study is the reliance on interpretable models to estimate LLM ability from benchmarks. Benchmarks often have poorly defined interpretations of what their scores measure, resulting in ambiguities in the meaning of the estimated abilities. While such a limitation would typically be a critical concern in psychology, sociology, or educational sciences, it remains a common and convenient practice in computer science.

Therefore, exploring the detailed interpretation of these ability estimates, residual dependencies in the data, and the precise meaning of CRM parameters represents a promising direction for further research. For example, developing benchmarks with clearly defined constructs that they are intended to measure could significantly advance this field.